\documentclass{article}
\usepackage{log_2026}						% for camera-ready version
\usepackage{booktabs}						% professional-quality tables
\usepackage{multirow}						% tabular cells spanning multiple rows
\usepackage{amsfonts}						% blackboard math symbols
\usepackage{graphicx}						% figures
\usepackage{duckuments}						% sample images

\usepackage[numbers,compress,sort]{natbib}	% for numerical citations
\usepackage{newunicodechar}

\newunicodechar{∈}{\ensuremath{\in}}
\newunicodechar{≈}{\ensuremath{\approx}}
\newunicodechar{⇒}{\ensuremath{\Rightarrow}}
\newunicodechar{ℤ}{\ensuremath{\mathbb{Z}}}
\newunicodechar{⊆}{\ensuremath{\subseteq}}

\usepackage{caption}
\usepackage{multirow}
\usepackage{amsmath}
\usepackage{subfiles}
\usepackage{array}
\usepackage{graphicx}
\usepackage{subfigure}
\usepackage{amsthm}
\usepackage{wrapfig}
\usepackage{booktabs} 
\usepackage{subcaption}
\usepackage{makecell}
\usepackage{adjustbox}
\usepackage{wrapfig}
\usepackage{multirow}

\usepackage{tikz}
\usetikzlibrary{positioning,arrows.meta}
\usepackage{xcolor}

\usepackage{float}

\usepackage{fancyvrb}
\usepackage{caption}

\usepackage{fancyvrb}
\usepackage{fvextra}   % <-- THIS is the missing piece

\usepackage{tcolorbox}
\tcbuselibrary{skins,breakable,listings} % listings instead of raw verbatim
\usepackage{courier} % monospaced font for listing

\usepackage{algorithm}
\usepackage{algpseudocode}

\title{Traceable Context-Enriched Knowledge Graph Generation with Induced Schemas}

\author[Abolhasani et al.]{%
Mohammad Sadeq Abolhasani\thanks{Corresponding author.}\\
Arizona State University\\
\email{mabolhas@asu.edu}\And
Yang Ba\\
Arizona State University\\
\email{yangba@asu.edu}\And
Yixuan He\\
Arizona State University\\
\email{Yixuan.He@asu.edu}\And
Rong Pan\\
Arizona State University\\
\email{Rong.Pan@asu.edu}
}

\begin{document}

\maketitle

\begin{abstract}
Knowledge graph generation typically relies either on predefined ontologies or on schema-free extraction. Ontology-driven pipelines enforce consistent typing but require costly schema design and maintenance, whereas schema-free methods often produce fragmented graphs with weak global organization, especially in long technical documents with dense, context-dependent information. We propose \textbf{TRACE-KG} (Text-dRiven schemA for Context-Enriched Knowledge Graphs), a framework that jointly constructs a context-enriched knowledge graph and an induced schema without assuming a predefined ontology. TRACE-KG captures conditional relations through structured qualifiers and organizes entities and relations using a data-driven schema that serves as a reusable semantic scaffold while preserving full traceability to the source evidence. Experiments show that TRACE-KG produces structurally coherent, traceable knowledge graphs and offers a practical alternative to both ontology-driven and schema-free construction pipelines. The code, experiments and prompts are available at \url{https://github.com/mosadddeq01/TRACE-KG}.

\end{abstract}

\section{Introduction}
% \section{Introduction}

Knowledge graphs (KGs) are increasingly used as long-lived substrates for organizing, integrating, and querying knowledge across documents, systems, and tasks~\citep{hogan2021knowledge}. 
Unlike raw text or traditional data models, graphs make structure explicit by representing entities and multi-relational links directly, supporting path-based queries and enabling \emph{gradual schema commitment}: a graph can begin with a lightweight structure and evolve as requirements change rather than requiring a fixed schema from the outset. 
At the same time, large language models (LLMs)~\citep{naveed2025comprehensive} have enabled powerful text-centric workflows such as retrieval-augmented generation (RAG)~\citep{lewis2020retrieval}, which can answer many \emph{local} questions directly from unstructured corpora.

Recent work suggests that explicit structured knowledge becomes particularly valuable when queries are \emph{global} rather than local—when answers must integrate evidence across documents, enforce consistency, or remain inspectable and consistently organized across tasks and time~\citep{khorashadizadeh2024research,bian2025llm}. In such settings, a KG acts not merely as an index over text but as a persistent structured representation layer that complements LLMs. Yet constructing high-quality KGs directly from complex technical documents remains challenging and largely unsolved~\citep{xue2022knowledge,zhong2023comprehensive}. A key reason is that useful KGs require more than extracting plausible triples: they require \emph{corpus-level organization}, typically provided by a \emph{schema}---a shared semantic vocabulary of entity and relation types that promotes normalization and more consistent organization across documents~\citep{hogan2021knowledge}. However, most existing text-to-KG pipelines operate in one of two common extremes.

At one extreme, schema-free approaches extract entities and relations directly from text without enforcing a shared type vocabulary or corpus-level consolidation~\citep{mo2025kggen,mihindukulasooriya2023text2kgbench}. While this often yields high \emph{local recall}, the resulting graphs fragment globally: the same entity may appear under multiple lexical variants, and semantically similar relations proliferate as near-duplicate predicates, in extreme cases producing relation vocabularies nearly as large as the edge set~\citep{hofer2024construction,bian2025llm,mo2025kggen}. At the opposite extreme, ontology-driven pipelines map text into a predefined ontology with fixed entity and relation categories~\citep{mihindukulasooriya2023text2kgbench,hofer2024construction}. Although this enforces consistent typing and semantics~\citep{hogan2021knowledge}, it assumes a suitable ontology exists and remains aligned with evolving terminology and document granularity. In practice, ontology construction and maintenance require substantial expert effort and often become a bottleneck~\citep{hofer2024construction,abolhasani2024leveraging}. Moreover, restricting extraction to a fixed ontology can exclude information that does not map cleanly to predefined classes or relations, while relation vocabularies often remain comparatively coarse even when entity typing is enforced~\citep{mo2025kggen,bai2025autoschemakg}.

These limitations are particularly acute in complex domain-specific corpora, where terminology is dense, lexical variation is high, external reference knowledge may be unavailable, and critical information is distributed across narrative text as well as figures, tables, diagrams, and equations~\citep{sun2025lkd,bian2025llm}. We therefore use a corpus of engineering and maintenance technical documents as a motivating case study, highlighting key challenges for text-to-KG pipelines, including consolidation, schema induction, and context handling. A central difficulty in such settings is the interplay between \emph{lexical} and \emph{semantic} heterogeneity~\citep{hofer2024construction,chen2024entity}: the same concept may appear under multiple surface forms, while similar expressions may denote different concepts depending on context. Unresolved lexical variation fragments evidence across duplicate nodes, whereas over-merging semantically distinct entities introduces incorrect relations~\citep{wang-etal-2025-aelc,pons2024knowledge,ding2024entgpt,wang2025match}. Moreover, relations in technical documents are often conditional---holding only under specific operating modes, temporal intervals, or constraints---so encoding them as unconditional triples can produce oversimplified or contradictory graphs~\citep{hogan2021knowledge,jiang2019role,xu2024context}. Capturing contextual qualifiers is therefore essential for faithful and inspectable knowledge representations.

These observations motivate a third approach between ontology-driven and
schema-free settings. Ontology-driven pipelines assume a predefined ontology,
$(D,O)\rightarrow {G_{\mathrm{on}}}$, whereas schema-free
pipelines extract a graph directly, $D\rightarrow G_{\mathrm{sf}}$.
To address the gap between these settings, we introduce \textbf{TRACE-KG}
(\textbf{T}ext-d\textbf{R}iven schem\textbf{A} for
\textbf{C}ontext-\textbf{E}nriched \textbf{K}nowledge \textbf{G}raphs),
a data-driven framework for \emph{schema-inducing KG generation}: given only
a corpus $D$, TRACE-KG constructs a context-enriched graph $G=(E,R,S)$,
where $E$ contains resolved entities, $R$ contains grounded relation
instances, and $S$ is an induced schema over both entities and relations.
TRACE-KG supports multimodal document ingestion, iterative entity and relation
resolution, relation canonicalization, structured qualifiers for conditional
relations, and source-grounded traceability. Our main contributions are as
follows:

\begin{itemize}
\item We introduce \textbf{TRACE-KG}, an end-to-end framework for constructing \textbf{context-enriched} knowledge graphs from complex documents, supporting multimodal document ingestion and integrating qualifier-preserving relation representation with provenance-grounded, schema-inducing KG construction.

% \item We introduce \textbf{TRACE-KG}, a multimodal, end-to-end framework for constructing \textbf{context-enriched} knowledge graphs from complex documents, with \textbf{explicit modeling of condition-aware relations} and provenance-preserving traceability.

\item We propose \textbf{TRACE-Schema}, a data-driven schema induction mechanism that organizes \textbf{both} entities and relations into a \textbf{reusable semantic scaffold}, enabling corpus-level consolidation without relying on a predefined ontology.

\item We develop an \textbf{evaluation framework} that compares the induced schema with human-created ontologies, enabling systematic evaluation of \textbf{schema coverage}, \textbf{granularity alignment}, and \textbf{structural consistency}.

\end{itemize}

\section{Related Work}

% % =========================
% % Main paper (short version)
% % =========================
% % \section{Related Work}

\paragraph{Schema commitment in text-to-KG construction.}
A central distinction in LLM-based text-to-KG pipelines is \emph{when} the schema is fixed. Ontology-driven approaches populate a predefined conceptual model, improving normalization and semantic consistency when the ontology matches the corpus \citep{mihindukulasooriya2023text2kgbench,ameri6017381tabular,abolhasani2024leveraging,tahsin2024generation}, but they depend on costly ontology design and are brittle under evolving terminology and corpus-specific granularity \citep{hofer2024construction}. At the other extreme, schema-free pipelines avoid predefined ontologies but often fragment entity and relation vocabularies. KGGen addresses this via post-hoc clustering and LLM-guided consolidation \citep{mo2025kggen}, and more recent work moves toward induced schema \citep{bai2025autoschemakg,sun2025lkd}; however, such schemas remain partial, uneven across entities and relations, and weakly coupled to corpus-level consolidation.

\paragraph{Local resolution, context, and reliability.}
Another line of work focuses on improving local grounding and extraction quality. EntGPT and comparison-based prompting improve entity linking and matching by constraining LLM decisions \citep{ding2024entgpt,wang2025match}, KG-assisted disambiguation exploits existing graph structure \citep{pons2024knowledge}, and schema-conditioned extractors such as GLiNER2 support structured extraction under user-provided label sets \citep{zaratiana2025gliner2}. Separately, prior work shows that many scientific and technical relations are conditional rather than unconditional triples, motivating qualifier- or context-aware representations \citep{jiang2019role,xu2024context,qin2025semantic}, and robustness-oriented workflows such as KARMA \citep{lu2025karma}. Wikontic~\cite{chepurova2026wikontic} likewise incorporates qualifiers in ontology-aware KG construction, demonstrating that qualifier modeling itself is not unique to TRACE-KG. TRACE-KG instead integrates qualifier-preserving relations with corpus-induced entity and relation schemas and source-grounded provenance without assuming a predefined ontology.

The core challenge is not extraction quality in isolation, but open-world \emph{co-construction} of a graph and its schema: a reusable schema over both entities and relations must be induced from the corpus while entity identity, relation canonicalization, conditional context, and provenance remain jointly consistent across the entire graph. Extended discussion appears in Appendix~\ref{app:extended_related_work}.

\section{Preliminaries}
\subsection{Corpus and text units}

Let $D=\{d_1,\dots,d_N\}$ denote a corpus of documents. Each document is converted into a provenance-preserving text stream, where textualized descriptions of non-text elements such as figures, tables, diagrams, or equations may be integrated when available. The stream is partitioned into sentence-preserving chunks. Let $C$ denote the set of sentence-preserving chunks. Each chunk $c\in C$ has an identifier $\mathrm{id}(c)$, textual content $\mathrm{text}(c)$, and provenance metadata $\mathrm{prov}(c)$ linking it to its source region.

\subsection{Entities and relations}

An \emph{entity mention} is a text span $m$ in a chunk $c$ that refers to an object, event, process, or abstract concept. Let $M$ denote the set of extracted mentions, where each mention stores its source chunk $\mathrm{chunk}(m)\in C$, span, name, description, type hint, confidence score, and justification excerpt. For a chunk $c$, let $M(c)=\{m\in M:\mathrm{chunk}(m)=c\}$.

A \emph{resolved entity} is a canonical graph-level object obtained by consolidating mentions that refer to the same underlying object or concept. Let $E$ denote the set of resolved entities, with $\mathrm{ResolvedEnt}:M\rightarrow E$ mapping each mention to its resolved entity. The entities supported by chunk $c$ are $E(c)=\{\mathrm{ResolvedEnt}(m):m\in M(c)\}$. Each entity $e\in E$ may also carry intrinsic properties $\mathrm{intr}(e)$, represented as typed key--value annotations.

Let $R$ denote the set of relation instances. A relation instance $r\in R$ is a directed, source-grounded statement
$r=(e_s,\ell_{\mathrm{raw}},e_t,Q,\pi,s)$,
where $e_s,e_t\in E$ are the source and target entities, $\ell_{\mathrm{raw}}$ is the raw predicate label, $Q$ is a set of contextual qualifiers, $\pi$ is provenance metadata, and $s\in[0,1]$ is a confidence score. Multiple relation instances may connect the same ordered entity pair, so the graph is represented as a directed multigraph. We denote the set of canonical relation labels by $\mathrm{RelCan}$ and use $\mathrm{CanonicalRel}:R\rightarrow \mathrm{RelCan}$ to assign each relation instance to a canonical label. Qualifiers $Q$ encode contextual conditions such as temporal, spatial, operational, conditional, uncertainty, or causal constraints.

\subsection{Induced schema and context-enriched graph}

TRACE-KG organizes entities and relations through an induced schema. In this paper, a \emph{schema} denotes a data-driven vocabulary and hierarchy of entity and relation types used to organize the generated graph.

At the entity level, $\mathrm{EntCls}$ denotes induced entity classes and $\mathrm{EntClsGroup}$ denotes broader entity class groups, with mappings $\tau_{\mathrm{ent}}:E\rightarrow \mathrm{EntCls}$ and $\gamma_{\mathrm{ent}}:\mathrm{EntCls}\rightarrow \mathrm{EntClsGroup}$.

At the relation level, $\mathrm{RelCls}$ denotes induced relation classes and $\mathrm{RelClsGroup}$ denotes broader relation class groups, with mappings $\tau_{\mathrm{rel}}:\mathrm{RelCan}\rightarrow \mathrm{RelCls}$ and $\gamma_{\mathrm{rel}}:\mathrm{RelCls}\rightarrow \mathrm{RelClsGroup}$.

Together, these define the induced schema
\(S=(\mathrm{EntCls},\allowbreak\mathrm{EntClsGroup},\allowbreak\mathrm{RelCls},\allowbreak\mathrm{RelClsGroup},\allowbreak\tau_{\mathrm{ent}},\allowbreak\gamma_{\mathrm{ent}},\allowbreak\tau_{\mathrm{rel}},\allowbreak\gamma_{\mathrm{rel}})\).

A \emph{context-enriched knowledge graph} is defined as $G=(E,R,S)$, where entities are associated with semantic descriptions, intrinsic properties, schema assignments, confidence estimates, justification excerpts, and provenance links, while relation instances additionally carry raw and canonical predicate labels together with contextual qualifiers. Intrinsic properties and qualifiers are represented as annotations attached to nodes and edges rather than as separate graph elements. Provenance links connect entities and relations to their supporting source evidence, enabling source-grounded traceability and auditability.

\section{Methodology}

\label{sec:method}
Given a corpus $D$, TRACE-KG constructs a context-enriched knowledge graph
$G=(E,R,S)$ grounded in document evidence and organized through an induced schema.
The framework is built around a repeated three-stage pattern: (i) high-recall recognition, which extracts local candidates while deferring global normalization; (ii) semantic neighborhood formation, which embeds multi-field representations and clusters them into candidate neighborhoods; and (iii) constrained resolution, where an LLM selects structured actions from the stage-specific action set $\mathcal{A}$ over explicit identifiers and deterministic validators execute and log the graph updates.
The LLM therefore proposes semantic decisions but does not directly mutate the graph state.
Algorithm~\ref{alg:tracekg_resolution} abstracts this mechanism for EntRes, EntClsRes, and RelRes, while Figure~\ref{fig:tracekg_overview} shows the full workflow.
Implementation details are provided in Appendix~\ref{app:method_details}.

% \begin{algorithm}[t]
% \captionsetup{font=footnotesize,skip=2pt}
% \footnotesize
% \caption{TRACE-KG constrained resolution pattern}
% \label{alg:tracekg_resolution}
% \begin{algorithmic}[1]
% \Require Candidate objects $X$; representation function $\phi$; action set $\mathcal{A}$; budget $T$
% \Ensure Refined objects $X'$ with logged edits
% \For{$t=1,\dots,T$}
%     \State Embed $X$ using $\phi$ and cluster into neighborhoods $\mathcal{B}$
%     \ForAll{$B\in\mathcal{B}$}
%         \State Select structured LLM actions from $\mathcal{A}$ over IDs in $B$
%         \State Validate identifiers, action format, and operation constraints
%         \State Execute approved actions deterministically and log edits
%     \EndFor
%     \State Refresh representations after applied edits
%     % \If{edits plateau} \textbf{break} \EndIf
%     \textcolor{orange}{ \If{stage-specific stopping criterion is met} \State \textbf{break} \EndIf }
% \EndFor
% \State \Return refined candidates $X'$
% \end{algorithmic}
% \end{algorithm}

\begin{algorithm}[t]
\captionsetup{font=footnotesize,skip=1pt}
\footnotesize
\caption{TRACE-KG constrained resolution pattern}
\label{alg:tracekg_resolution}

\begin{algorithmic}[1]
\Require Candidates $X$; representation $\phi$; actions $\mathcal{A}$; budget $T$
\Ensure Refined $X'$ with logged edits

\For{$t=1,\dots,T$}
    \State Embed $X$ with $\phi$ and cluster into neighborhoods $\mathcal{B}$
    \ForAll{$B\in\mathcal{B}$}
        \State Select structured LLM actions from $\mathcal{A}$ over IDs in $B$
        \State Validate identifiers, format, and operation constraints
        \State Execute approved actions deterministically and log edits
    \EndFor
    \State Refresh representations after applied edits
    \State \textbf{if} stage-specific stopping criterion is met \textbf{then break}
\EndFor
\State \Return $X'$
\end{algorithmic}

\end{algorithm}

\begin{figure*}[t]
    \centering
    \includegraphics[width=\linewidth]{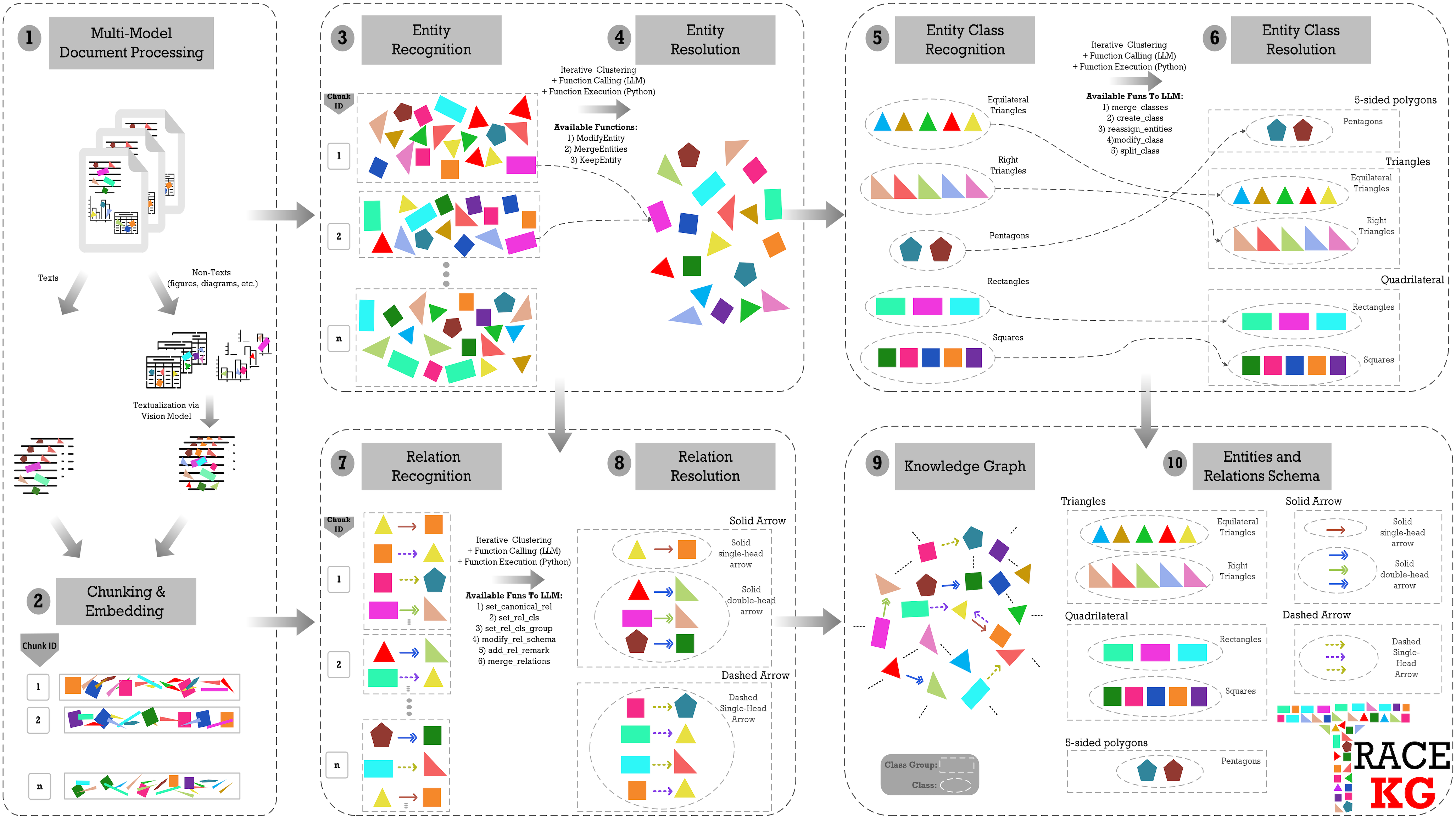}
    \captionsetup{font=scriptsize,skip=2pt}
    \caption{TRACE-KG pipeline. 
    (1--2) \textbf{Processing/Chunking}: documents are textualized and chunked (scattered, unorganized shapes). 
    (3) \textbf{EntRec}: local entity mentions are extracted but remain unresolved (e.g., duplicate pink rectangles). 
    (4) \textbf{EntRes}: duplicate mentions are consolidated into resolved entities (e.g., one pink rectangle). 
    (5) \textbf{EntClsRec}: resolved entities are grouped into entity classes (e.g., rectangles grouped together). 
    (6) \textbf{EntClsRes}: entity classes are refined into broader class groups (e.g., rectangles and squares under a broader quadrilateral group). 
    (7) \textbf{RelRec}: grounded relation instances are extracted with qualifiers (e.g., mixed solid single-headed, solid double-headed, and dashed single-headed arrows). 
    (8) \textbf{RelRes}: relation labels are canonicalized and organized into relation classes and class groups (e.g., arrow types separated into solid single-headed, solid double-headed, and dashed single-headed arrows, then grouped into broader solid and dashed relation groups). 
    (9) \textbf{KG}: resolved entities and relation instances form the context-enriched KG. 
    (10) \textbf{Schema}: induced entity and relation hierarchies form the schema. 
    Stages (4), (6), and (8) use semantic clustering, structured LLM actions, and deterministic execution.}
    \label{fig:tracekg_overview}
\end{figure*}

\subsection{Document ingestion and chunking}
\label{sec:chunking}

% Documents are converted into unified, provenance-preserving text streams. When non-text regions such as figures, tables, diagrams, or equations are available, TRACE-KG supports textualizing them with a vision-language document processor while preserving page- and region-level provenance. The resulting streams are segmented into sentence-preserving chunks $c\in C$. Later stages may use neighboring chunks for disambiguation, but extracted entities and relations must be grounded in the focus chunk.

Documents are converted into unified, provenance-preserving text streams. When non-text regions such as figures, tables, diagrams, or equations are available, TRACE-KG supports multimodal ingestion by textualizing these regions with a vision-language document processor while preserving page- and region-level provenance. The resulting textual evidence is processed by the same downstream KG construction pipeline. A text-only versus text+VLM ablation in Appendix~\ref{app:casestudy} evaluates the contribution of this ingestion step. The resulting streams are segmented into sentence-preserving chunks $c \in C$. Later stages may use neighboring chunks for disambiguation, but extracted entities and relations must be grounded in the focus chunk.

\subsection{Entity layer: recognition and iterative resolution}
\label{sec:entity_layer}

\paragraph{Entity recognition (EntRec).}
EntRec performs high-recall, chunk-local mention extraction. For each focus chunk $c$, the extractor receives $\mathrm{text}(c)$ and a short window of preceding chunks for disambiguation, but extracts mentions only from the focus chunk. Each mention includes a mention-level name, description, broad type hint, confidence score, and justification excerpt grounded in $\mathrm{text}(c)$. Explicit intrinsic properties are extracted as typed key--value annotations when supported by the text. At this stage, TRACE-KG does not impose a fixed schema; mentions are collected before global identity and class decisions are made.

\paragraph{Entity resolution (EntRes).}
EntRes consolidates mention-level candidates into resolved entities. Mentions are represented using multiple fields, including name, description, type hint, justification evidence, and resolution context, then clustered with HDBSCAN into candidate identity neighborhoods. Within each neighborhood, the LLM selects structured actions $\mathcal{A}$ over provided identifiers, such as \textsc{MergeEntities}, \textsc{ModifyEntity}, or \textsc{KeepEntity}. The graph state is updated only after deterministic validators check identifier validity, action format, and operation constraints. EntRes follows Algorithm~\ref{alg:tracekg_resolution}: after each round, mention-to-entity assignments and multi-field representations are refreshed, neighborhoods are recomputed, and the process stops when a round produces zero accepted
merge actions or after five iterations.
% the process stops when merge activity plateaus or the iteration budget is reached. 
This keeps identity edits auditable while allowing local resolution decisions to propagate across later neighborhoods.

\subsection{Entity schema induction: classes and class groups}
\label{sec:entity_schema}

After entity resolution, TRACE-Schema organizes resolved entities into reusable schema-level types without requiring a predefined ontology.

\paragraph{Entity class recognition (EntClsRec).}
EntClsRec embeds each resolved entity using its name, description, intrinsic properties, broad type hint, and representative supporting evidence. These representations are clustered into semantic neighborhoods, and the LLM proposes candidate entity classes with labels, descriptions, and member entity identifiers. A neighborhood may yield multiple class proposals when it contains heterogeneous entities, avoiding forced assignment to an overly broad class. A fallback pass handles entities not confidently covered by earlier proposals.

\paragraph{Entity class resolution (EntClsRes).}
EntClsRes refines candidate classes into a validated schema hierarchy. Candidate class representations are embedded and clustered, then resolved through Algorithm~\ref{alg:tracekg_resolution} using structured actions such as \textsc{MergeClasses}, \textsc{SplitClass}, \textsc{CreateClass}, \textsc{ReassignEntities}, and \textsc{ModifyClass}. The output assigns each resolved entity to an entity class and broader class group, yielding
$\mathrm{EntClsGroup}\rightarrow \mathrm{EntCls}\rightarrow E$.

\subsection{Relation layer: chunk-centric recognition with qualifiers}
\label{sec:relation_recognition}

\paragraph{Relation recognition (RelRec).}
RelRec extracts grounded relation instances after entities have been resolved and assigned to entity classes. For each chunk $c$, the extractor receives $\mathrm{text}(c)$ and the resolved entities supported by that chunk, $E(c)$. It may output only directed relations whose endpoints are in $E(c)$ and whose evidence appears in the focus chunk. Each relation instance records subject and object identifiers, a raw predicate label, an instance description, confidence, justification evidence, provenance, and a qualifier dictionary.
Qualifiers capture the local validity context of a relation, including temporal, spatial, operational, conditional, uncertainty, or causal constraints. They are attached as edge-level annotations rather than converted into additional graph nodes or edges. This preserves \emph{when}, \emph{where}, or \emph{under what conditions} a relation holds while keeping the graph topology focused on entity--entity relations.

\subsection{Relation canonicalization and relation schema induction}
\label{sec:relation_schema}

Relation processing follows the same recognition--neighborhood--resolution pattern as the entity layer, but its effect on the graph differs. Entity resolution collapses duplicate mentions into shared nodes. In contrast, relation instances are preserved as grounded evidence because each instance is tied to a specific source context and ordered entity pair. Therefore, relation resolution primarily normalizes predicate semantics and assigns schema annotations rather than deleting edges.

\paragraph{Relation resolution (RelRes).}
RelRes embeds each relation instance using a multi-field representation combining the raw predicate label, relation description, endpoint names, endpoint schema metadata, qualifiers, and supporting evidence. The resulting neighborhoods are resolved through Algorithm~\ref{alg:tracekg_resolution} using structured actions such as \textsc{SetCanonicalRel}, \textsc{SetRelCls}, \textsc{SetRelClsGroup}, \textsc{ModifyRelSchema}, \textsc{AddRelRemark}, and \textsc{MergeRelations}. The action \textsc{MergeRelations} is permitted only when relation instances connect the same ordered entity pair and express equivalent semantics; otherwise, instances are retained and only their canonical labels or schema assignments are updated.

% After each round, canonical relation labels and schema assignments are refreshed, relation representations are recomputed, and refinement continues until edits plateau or the iteration budget is reached. 
RelRes currently performs a single pass over the relation clusters, with all proposed actions validated before deterministic execution. The output is a set of grounded relation instances $R$, a canonical relation vocabulary $\mathrm{RelCan}$, and the relation schema hierarchy
$\mathrm{RelClsGroup}\rightarrow \mathrm{RelCls}\rightarrow \mathrm{RelCan}$.
The relation hierarchy supports abstraction at multiple levels: canonical relations preserve fine-grained predicate semantics, relation classes group semantically related predicates, and relation class groups provide broader categories for reuse and querying. This allows queries to operate over a higher-level relation category without enumerating every canonical predicate, while retaining the more specific relation labels and grounded instances for inspection.

% \begin{table*}[t]
% \caption{Knowledge retention results on MINE-1, grouped by construction setting.}
% \label{tab:mine1_main}

% \centering
% \small
% \vspace{0.4em}
% \begingroup
% \setlength{\tabcolsep}{1pt}
% \renewcommand{\arraystretch}{1.18}

% \begin{tabular*}{0.98\textwidth}{@{\extracolsep{\fill}}lcccccccc}
% \toprule
% Method 
% & Ret.Acc 
% & EGU$\uparrow$ 
% & RWA$\uparrow$ 
% & SCI$\uparrow$ 
% & Leak$\downarrow$ 
% & TriCR$\to 1$ 
% & Conn.$\uparrow$ 
% & AvgRank$\downarrow$ \\
% \midrule

% \multicolumn{9}{l}{\textbf{Schema-free extraction / post-hoc consolidation}} \\
% OpenIE        
% & 56.0\% 
% & 40.5\%          
% & 41.5\%          
% & 0.024          
% & 2.3\%  
% & 3.201          
% & 74.0\% 
% & 3.56 \\

% KGGen         
% & 63.6\% 
% & 30.1\%          
% & 30.1\%          
% & 0.005          
% & \textbf{0.0\%}  
% & 0.494          
% & 46.1\% 
% & 4.00 \\

% \midrule
% \multicolumn{9}{l}{\textbf{Retrieval-oriented graph baseline}} \\
% GraphRAG      
% & 48.4\% 
% & 45.1\%          
% & 45.1\%          
% & 0.146          
% & \textbf{0.0\%}  
% & 0.394          
% & \textbf{91.5\%} 
% & 2.67 \\

% \midrule
% \multicolumn{9}{l}{\textbf{Dynamic schema induction baseline}} \\
% AutoSchemaKG  
% & \textbf{95.1\%} 
% & 37.3\% 
% & 58.7\%          
% & 0.047          
% & 36.5\% 
% & 3.599          
% & 61.5\% 
% & 3.33 \\

% \midrule
% \multicolumn{9}{l}{\textbf{Proposed schema-inducing context-enriched construction}} \\
% TRACE-KG      
% & 90.2\% 
% & \textbf{78.5\%} 
% & \textbf{79.6\%} 
% & \textbf{0.270} 
% & 1.3\%  
% & \textbf{0.956} 
% & 88.5\% 
% & \textbf{1.44} \\

% \bottomrule
% \end{tabular*}

% \endgroup
% \end{table*}

\subsection{Graph assembly and audit controls}
\label{sec:kg_assembly}

The final stage assembles the context-enriched knowledge graph $G=(E,R,S)$. Resolved entities become nodes annotated with descriptions, intrinsic properties, schema assignments, confidence scores, justification excerpts, and provenance links. Relation instances become directed edges annotated with raw and canonical predicate labels, relation-schema assignments, qualifier dictionaries, confidence scores, justification excerpts, and provenance.

TRACE-KG applies audit controls throughout construction rather than treating LLM outputs as direct graph updates. Entity and relation extraction are chunk-local and must be grounded in focus-chunk evidence. Resolution is restricted to candidate neighborhoods, limiting unconstrained global edits. LLM decisions are expressed as structured actions over existing identifiers, while deterministic validators execute the approved updates and log the resulting edits. These controls do not guarantee correctness, but they make generated nodes, edges, schema assignments, and resolution decisions inspectable against their supporting evidence.
\label{sec:method}

\section{Experiment \& Analysis}

% \section{Experiment \& Analysis}

% TRACE-KG is evaluated along two complementary axes: 
% (i) whether the constructed graph supports reliable graph-based retrieval of factual evidence, and 
% (ii) whether the induced schema recovers reusable entity and relation types without access to a predefined ontology. 
% Implementation details appear in the appendix, and an anonymized codebase with full experiment scripts is available at \url{https://anonymous.4open.science/r/TRACE-KG}.

TRACE-KG is evaluated along three complementary axes: (i) whether the constructed graph supports reliable graph-based retrieval of factual evidence, (ii) whether the induced schema recovers reusable entity and relation types without access to a predefined ontology, and (iii) whether joint processing of multiple technical documents supports cross-document consolidation and schema reuse. Implementation details appear in the appendix, and the codebase with full experiment scripts and prompts are available at \url{https://github.com/mosadddeq01/TRACE-KG}.

% ============================================================
% Experiment 1 (Main paper): Knowledge Retention on MINE-1
% ============================================================

\begin{table*}[t]
\caption{Knowledge retention results on MINE-1, grouped by construction setting.}
\label{tab:mine1_main}

\centering
\small
\vspace{0.4em}
\begingroup
\setlength{\tabcolsep}{1pt}
\renewcommand{\arraystretch}{1.18}

\begin{tabular*}{0.98\textwidth}{@{\extracolsep{\fill}}lcccccccc}
\toprule
Method 
& Ret.Acc 
& EGU$\uparrow$ 
& RWA$\uparrow$ 
& SCI$\uparrow$ 
& Leak$\downarrow$ 
& TriCR$\to 1$ 
& Conn.$\uparrow$ 
& AvgRank$\downarrow$ \\
\midrule

\multicolumn{9}{l}{\textbf{Schema-free extraction / post-hoc consolidation}} \\
OpenIE        
& 56.0\% 
& 40.5\%          
& 41.5\%          
& 0.024          
& 2.3\%  
& 3.201          
& 74.0\% 
& 3.56 \\

KGGen         
& 63.6\% 
& 30.1\%          
& 30.1\%          
& 0.005          
& \textbf{0.0\%}  
& 0.494          
& 46.1\% 
& 4.00 \\

\midrule
\multicolumn{9}{l}{\textbf{Retrieval-oriented graph baseline}} \\
GraphRAG      
& 48.4\% 
& 45.1\%          
& 45.1\%          
& 0.146          
& \textbf{0.0\%}  
& 0.394          
& \textbf{91.5\%} 
& 2.67 \\

\midrule
\multicolumn{9}{l}{\textbf{Dynamic schema induction baseline}} \\
AutoSchemaKG  
& \textbf{95.1\%} 
& 37.3\% 
& 58.7\%          
& 0.047          
& 36.5\% 
& 3.599          
& 61.5\% 
& 3.33 \\

\midrule
\multicolumn{9}{l}{\textbf{Proposed schema-inducing context-enriched construction}} \\
TRACE-KG      
& 90.2\% 
& \textbf{78.5\%} 
& \textbf{79.6\%} 
& \textbf{0.270} 
& 1.3\%  
& \textbf{0.956} 
& 88.5\% 
& \textbf{1.44} \\

\bottomrule
\end{tabular*}

\endgroup
\end{table*}

\subsection{Experiment 1: Knowledge Retention on MINE-1}
\label{sec:exp1_MINE1}

To assess factual knowledge retention under graph-based retrieval, we adopt the MINE-1 benchmark \cite{mo2025kggen}. Given a source article and a set of derived factual statements, the task is to determine whether each statement can be supported by a retrieved subgraph of the constructed graph. We compare TRACE-KG against four baselines representing different construction settings: schema-free extraction or post-hoc consolidation (OpenIE \cite{angeli2015leveraging}, KGGen \cite{mo2025kggen}), a retrieval-oriented graph baseline (GraphRAG \cite{larson2024graphrag}), and dynamic schema induction (AutoSchemaKG \cite{bai2025autoschemakg}). All methods are evaluated on identical benchmark instances.

\paragraph{Protocol (retrieval and strict judging).}
We follow the retrieval-and-judge procedure of MINE-1 \cite{mo2025kggen}. Each factual statement is embedded and matched to KG entities; the top-$k$ entities are retrieved, expanded within a fixed hop budget to form an induced subgraph, and evaluated by a strict LLM judge. The judge determines whether the statement is supported using \emph{only} the retrieved subgraph (binary decision; no external knowledge; no inferred edges). To ensure comparability, we use a fixed judge configuration across all methods, along with identical retrieval hyperparameters and embedding models.

\paragraph{Beyond raw retrieval accuracy.}
% The standard MINE-1 score (Ret.Acc) measures whether facts can be supported from retrieved subgraphs. However, high Ret.Acc can arise from undesirable artifacts, such as copying long text spans into entity strings or retrieving fragments from globally disconnected graphs. To better capture graph quality, we complement Ret.Acc with structural and representational metrics reflecting connectivity, compression, and information leakage. We report three composite metrics: \textbf{RWA} (Reachability-Weighted Accuracy), \textbf{EGU} (Effective Graph Utilization), and \textbf{SCI} (Structural Coherence Index):
The standard MINE-1 retrieval accuracy (Ret.Acc) is the primary task metric, measuring whether factual statements can be supported by retrieved subgraphs. We retain Ret.Acc as the primary measure and use the additional metrics below as complementary diagnostics rather than replacements for retrieval accuracy. High Ret.Acc can nevertheless arise from undesirable artifacts, such as copying long text spans into entity strings or retrieving fragments from globally disconnected graphs. To expose these failure modes, we complement Ret.Acc with structural and representational metrics reflecting connectivity, compression, and information leakage. We report three composite diagnostics: RWA (Reachability-Weighted Accuracy), EGU (Effective Graph Utilization), and SCI (Structural Coherence Index):
\begin{align}
\textbf{RWA} &= \text{Ret.Acc} \times \text{Conn.} \\
\textbf{EGU} &= \text{Ret.Acc} \times \text{Conn.} \times (1 - \text{Leak\%}) \\
\textbf{SCI} &= \text{AvgDeg} \times \text{Clust.} \times \text{Conn.}
\end{align}
% RWA adjusts retrieval accuracy by graph connectivity; EGU further penalizes lexical leakage; and SCI measures structural quality independently of retrieval accuracy (details in Appendix~\ref{sec:appendix_mine1}).

RWA adjusts retrieval accuracy by graph connectivity; EGU further penalizes
lexical leakage; and SCI measures structural quality independently of retrieval
accuracy. These quantities are intended to diagnose
connectivity, leakage, and structural coherence alongside Ret.Acc, not to
redefine the MINE-1 task objective (details in Appendix~\ref{sec:appendix_mine1}).

\paragraph{Results.}
% Table~\ref{tab:mine1_main} shows that Ret.Acc alone yields an incomplete ranking. Although AutoSchemaKG achieves the highest Ret.Acc (95.1\%), it also exhibits high leakage (36.5\%) and a large TriCR (3.599), indicating that much of the retained evidence remains close to the source text rather than compactly encoded in the graph. 
Table~\ref{tab:mine1_main} shows that TRACE-KG achieves high Ret.Acc (90.2\%), while AutoSchemaKG obtains the highest raw retrieval accuracy (95.1\%). The complementary diagnostics expose a different trade-off: AutoSchemaKG exhibits high leakage (36.5\%) and a large TriCR (3.599), indicating that much of the retained evidence remains close to the source text rather than compactly encoded in the graph. 
% TRACE-KG, by contrast, maintains high Ret.Acc (90.2\%) while achieving low leakage (1.3\%), a TriCR close to 1 (0.956), and strong connectivity (88.5\%), resulting in the best EGU, RWA, SCI, and AvgRank. 
TRACE-KG, by contrast, combines low leakage (1.3\%), a TriCR close to 1
(0.956), and strong connectivity (88.5\%), resulting in the best EGU, RWA,
SCI, and AvgRank.

\begin{figure*}[t]
\centering
\includegraphics[width=0.95\linewidth]{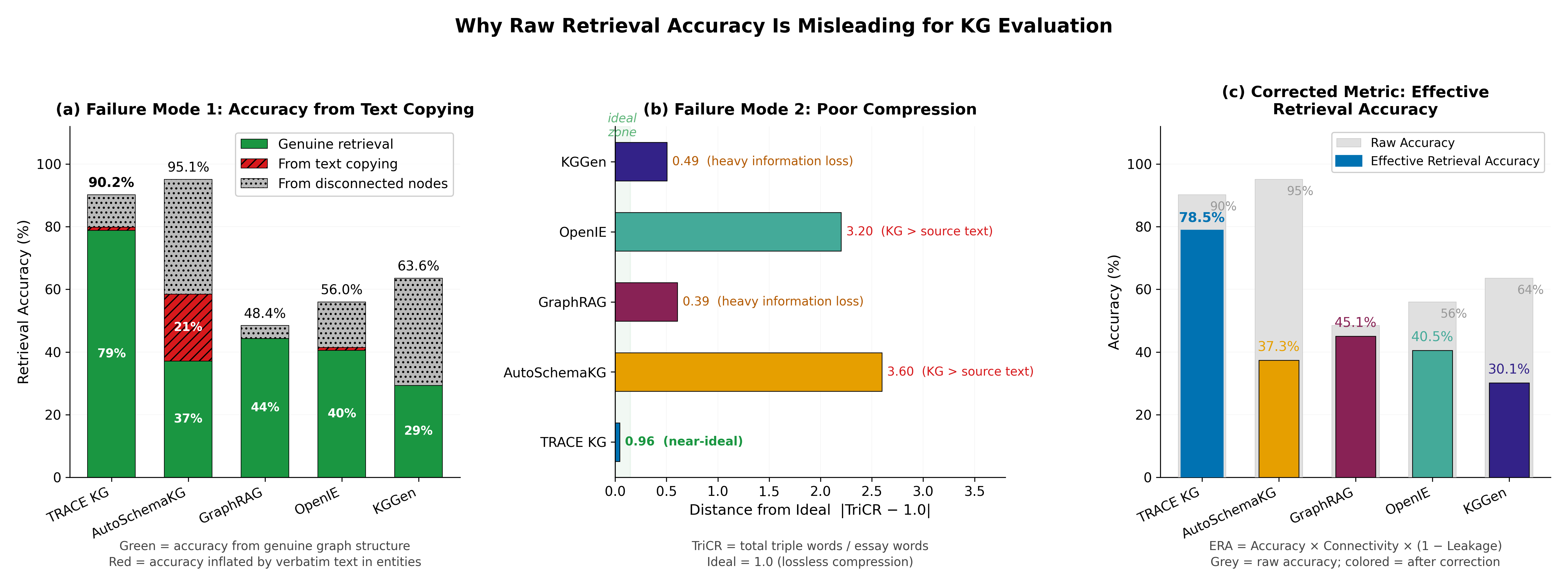}
\vspace{-6pt}
\caption{Failure modes behind raw retrieval accuracy on MINE-1. Left: Ret.Acc retained after discounting by connectivity and leakage. Middle: deviation of TriCR from the ideal value 1. Right: Ret.Acc versus EGU.}
\label{fig:mine1_failure_modes}
\vspace{-10pt}
\end{figure*}

These results show that TRACE-KG balances retrieval accuracy with structural coherence, rather than relying on lexical overlap. See Appendix~\ref{sec:appendix_mine1} for additional diagnostics and baseline analysis. Figure~\ref{fig:mine1_failure_modes} visualizes the effects of discounting leakage and structural fragmentation.

% final draft

% =========================================================
% Experiment 2 (Main paper)
% =========================================================

\subsection{Experiment 2: Ontology-Held-Out Schema Evaluation}
\label{sec:exp2_schema}

\paragraph{Goal and setting.}
We evaluate TRACE-KG’s ability to induce a reusable schema directly from text using a novel \emph{ontology-held-out} protocol. TRACE-KG operates solely on the input corpus and induces a hierarchical schema of entity and relation types as part of extraction and resolution, without access to any predefined ontology. A curated reference ontology is introduced only after schema induction for post hoc mapping and evaluation, ensuring that the induced schema is assessed against an independent, human-designed conceptual model (details in Appendix~\ref{app:exp2_schema_details}).

\paragraph{Why this dataset.}
This setting requires (i) human-curated ontologies as held-out ground truth and (ii) domain-specific text from which schemas must be induced. We therefore use the DBpedia-WebNLG domain ontologies and sentence splits distributed by Text2KGBench and OSKGC \cite{mihindukulasooriya2023text2kgbench, wang2025oskgc}. The dataset spans $19$ domains with diverse entity and relation vocabularies, enabling cross-domain evaluation of schema induction. Inputs consist of \emph{disjoint sentences} rather than full documents; thus, document-level structural cues (e.g., titles, sections) are absent, and schema must be inferred solely from local linguistic evidence.

\paragraph{Protocol.}
For each domain, we run TRACE-KG on the dataset-provided \texttt{train} sentences to induce a fixed TRACE schema. The induced schema contains hierarchical structures for both entities and relations: entity classes and broader entity class groups, and canonical relation labels, relation classes, and broader relation class groups. The reference ontology is then used only for post hoc alignment and evaluation. We report results under three \emph{evaluation scopes} defined by gold triples: \textbf{Source} (\texttt{train} gold triples), \textbf{Held-out} (\texttt{test} gold triples), and \textbf{Combined} (all gold triples). Alignment judgements are produced by a structured LLM verifier, with targeted human auditing of low-confidence cases.

\paragraph{Schema mapping and verification.}
To compare the induced TRACE schema with the held-out reference ontology, we retrieve the top-$K$ TRACE candidates for each active reference anchor and verify each pair using a structured verifier. Judgements are \textsc{Equivalent}, \textsc{Narrower}, \textsc{Broader}, or \textsc{Unrelated}; \textsc{Equivalent} counts as exact recovery and \textsc{Narrower} as compatible refinement. Relation alignment is direction-relaxed, low-confidence outputs are manually audited, and all candidates and judgements are retained for inspection.

\paragraph{Active anchors and metrics.}
For each evaluation scope (Source, Held-out, Combined), we evaluate only \emph{active} reference anchors, defined as ontology relations appearing in the gold triples together with their domain and range concepts. Anchors are frequency-weighted according to their occurrence in the corresponding split. We report coverage (Exact and Compatible), frequency-weighted MRR@5, and domain/range consistency (D/R).

\paragraph{Results and discussion.}
Table~\ref{tab:exp2_main} reports TRACE-KG's ontology-held-out schema evaluation, macro-averaged over 19 domains. Concept alignment is strong across all scopes, with near-saturated compatible coverage (97--99\%). Lower exact recovery ($\approx 50\%$) reflects frequent \textsc{Narrower} refinements rather than failure to recover the underlying concept. Relation alignment is more challenging, especially in the Held-out scope, but still reaches 84.8\% compatible coverage. Retrieval quality remains high overall, with Combined MRR@5 of 0.883 for concepts and 0.795 for relations. Aligned relations also show strong domain/range consistency, reaching 82.7\% under Held-out weighting. Overall, these results indicate that TRACE-KG induces schema elements directly from text that remain broadly compatible with a reference ontology.

\begin{table}[t]
\centering
\footnotesize

% =========================
% Table 2
% =========================
\begin{minipage}[t]{0.49\linewidth}
\vspace{0pt}
\centering

\captionof{table}{Ontology-held-out schema evaluation on
DBpedia-WebNLG (19 domains).}
\label{tab:exp2_main}

\setlength{\tabcolsep}{1.5pt}
\renewcommand{\arraystretch}{0.90}

\begin{tabular}{lccc}
\toprule
\textbf{Metric} & \textbf{Source} & \textbf{Held-out} & \textbf{Combined} \\
\midrule

\multicolumn{4}{l}{\textit{Active reference anchors (avg.\ per ontology)}} \\
Concepts  & 11.2 & 12.8 & 14.3 \\
Relations & 28.4 & 31.1 & 42.7 \\
\midrule

\multicolumn{4}{l}{\textit{Concept alignment}} \\
Coverage (Compat)   & 99.1\% & 97.2\% & 98.3\% \\
Coverage (Exact)    & 50.2\% & 51.2\% & 50.2\% \\
Coverage (Narrower) & 48.9\% & 46.0\% & 48.1\% \\
MRR@5               & 0.889  & 0.877  & 0.883  \\
\midrule

\multicolumn{4}{l}{\textit{Relation alignment}} \\
Coverage (Compat)   & 92.4\% & 84.8\% & 89.9\% \\
Coverage (Exact)    & 61.8\% & 49.7\% & 56.8\% \\
Coverage (Narrower) & 30.6\% & 35.1\% & 33.1\% \\
MRR@5               & 0.885  & 0.694  & 0.795  \\
\midrule

\multicolumn{4}{l}{\textit{Domain/range consistency}} \\
D/R (L1--L3)        & 91.0\% & 82.7\% & 88.1\% \\
\bottomrule
\end{tabular}

\end{minipage}
\hfill
% =========================
% Table 3
% =========================
\begin{minipage}[t]{0.49\linewidth}
\vspace{0pt}
\centering

\captionof{table}{Cross-document consolidation and schema reuse on the
ISO~55000-family corpus. Separate runs process each standard independently;
the combined run processes all three jointly.}
\label{tab:iso_cross_document}

\setlength{\tabcolsep}{2pt}
\renewcommand{\arraystretch}{0.90}

\begin{tabular}{lcc}
\toprule
\textbf{Metric} & \textbf{Sep.\ runs} & \textbf{Combined} \\
\midrule
Pages / chunks          & 83 / 339 & 83 / 339 \\
Entity mentions         & 638      & 638      \\
Raw relation candidates & 882      & 847      \\
Resolved entities       & 403      & 285      \\
Entity classes          & 124      & 64       \\
Canonical relations     & 229      & 156      \\
Relation classes        & 113      & 62       \\
\midrule
Ent.\ cls.\ in $\geq$2 docs & -- & 39 / 64 \\
Ent.\ cls.\ in all 3 docs   & -- & 18 / 64 \\
Rel.\ cls.\ in $\geq$2 docs & -- & 36 / 62 \\
Rel.\ cls.\ in all 3 docs   & -- & 16 / 62 \\
\bottomrule
\end{tabular}

\end{minipage}

\end{table}

\subsection{Cross-Document Consolidation and Schema Reuse}
\label{sec:iso_cross_document}

%%%%%%%%%%%%

To evaluate corpus-level consolidation, we process three related ISO 55000-family
technical standards (83 pages, 339 chunks), independently and jointly.
Table~\ref{tab:iso_cross_document} shows that joint processing produces a more
compact resolved vocabulary and substantial cross-document schema reuse.
Full protocol and audit details are provided in Appendix~\ref{app:crossdoc}.
\subsection{Ablation and Backbone Sensitivity}
\label{sec:ablation}

Component ablations and backbone-sensitivity experiments show that each tested
component contributes to Ret.Acc and that performance varies across LLM
backbones. Full ablation and backbone-sensitivity results are provided in Appendix~\ref{app:ablation_analysis}, with a manual faithfulness and hallucination audit in Appendix~\ref{app:faithfulness_audit}.

% \paragraph{Component ablation.}
% We ablate major TRACE-KG components on 25 of the 100 MINE-1 articles (375 fact checks), using identical retrieval parameters, embeddings, and judge across settings. This subset is slightly easier than the full benchmark, yielding a full-model Ret.Acc of 91.4\% compared with 90.2\% on the complete benchmark. Removing qualifier extraction causes the largest Ret.Acc drop, suggesting that qualifiers materially affect whether retrieved evidence preserves the context needed to support a fact. Entity resolution and relation canonicalization also contribute substantially by reducing entity fragmentation and predicate inconsistency. Schema induction has a smaller effect on MINE-1, consistent with the benchmark's emphasis on local factual support rather than corpus-level organization.

% \paragraph{Backbone sensitivity.}
% To assess the effect of LLM capability, we evaluate TRACE-KG with four backbones on the same 25 MINE-1 articles (375 fact checks), keeping the pipeline, retrieval parameters, embeddings, and judge fixed across settings. Performance improves substantially with stronger backbones, indicating that TRACE-KG is sensitive to the capability of the underlying LLM. At the same time, the pipeline operates across all four model families, showing that its construction procedure is not specific to GPT-5.

\section{Conclusion}

We introduced TRACE-KG, a framework for constructing context-enriched and
traceable knowledge graphs from complex technical documents while inducing a
reusable schema over both entities and relations without a predefined ontology.
TRACE-KG integrates multimodal document ingestion, qualifier-preserving
relations, provenance-grounded extraction, and constrained LLM-guided resolution
with deterministic execution. Across complementary evaluations, TRACE-KG
demonstrates strong knowledge retention and structural quality, ontology-held-out
schema recovery, and cross-document consolidation and schema reuse. These
results support data-driven schema induction as a practical alternative between
ontology-driven and schema-free KG construction.

% \section{Limitations and Future Work}
% \subfile{sections/7-Limitations_and_Future_Work}

% \subsection{Bibliographies}

% For natbib users:
\bibliographystyle{unsrtnat}
\bibliography{reference}
% For bibLaTeX users:
% \printbibliography

\clearpage

\appendix
\section{Extended Related Work}

% \appendix
% \section{Extended Related Work}
\label{app:extended_related_work}

% Appendix (extended related work version)

\paragraph{Ontology-driven construction and ontology population.}
A prominent direction treats the schema as an explicit \emph{input}, framing text-to-KG construction as ontology population. \textsc{Text2KGBench} evaluates this setting by measuring whether generated triples remain faithful to text while conforming to ontology constraints, showing that errors persist even when the target schema is fixed \citep{mihindukulasooriya2023text2kgbench}. In industrial and maintenance settings, Ameri et al.\ operationalize ontology-guided construction using a curated thesaurus and work-order ontology to map noisy descriptions into RDF representations aligned with domain concepts \citep{ameri6017381tabular}. OntoKGen similarly employs LLMs in an interactive workflow to extract and refine an ontology from technical documents before generating a schema-consistent KG \citep{abolhasani2024leveraging}. While these approaches improve normalization through schema alignment, they remain sensitive to ontology design, costly to maintain over time, and brittle under evolving terminology and corpus-specific granularity \citep{hofer2024construction}.

\paragraph{Schema-free extraction and post-hoc consolidation.}
A complementary direction performs schema-free extraction, producing entities and relations directly from text. Such approaches often yield fragmented vocabularies at corpus scale: semantically similar relations appear under many surface predicates, and entities are duplicated across abbreviations and near-synonyms. KGGen highlights this issue, showing that naive extraction can produce relation vocabularies nearly as large as the edge set, and addresses it through embedding-based clustering with LLM-guided consolidation of entities and relations \citep{mo2025kggen}. While this improves representational reuse, it does not induce \emph{schema connectivity}: in schema-free graphs, objects are related primarily through extracted edges, whereas type- or class-level structure could connect entities (and relations) via shared membership even in the absence of direct edges.

\paragraph{Inducing schema from text.}
To move beyond post-hoc label normalization, recent approaches treat schema as an \emph{output} of construction. AutoSchemaKG induces abstract concepts for entities and relations and uses these assignments as a learned schema layer \citep{bai2025autoschemakg}. This enables data-driven organization without a predefined ontology, but the resulting schema elements can remain verbose and heterogeneous, and their coupling with robust corpus-level identity resolution remains limited. LKD-KGC targets domain-specific repositories by modeling cross-document dependencies: it orders documents, incorporates retrieved context during summarization, and induces an entity schema (type names and definitions) to guide extraction \citep{sun2025lkd}. While this leverages repository structure, the resulting organization is often uneven across entities and relations, and extracted predicates and nodes remain close to natural-language expressions, limiting stable and reusable schema formation across long corpora.

\paragraph{Entity resolution, entity matching, and schema-aware IE}
Mention-level ambiguity remains a central bottleneck in KG construction. EntGPT frames entity linking as a structured generative workflow based on explicit candidate sets, showing that grounding decisions can materially improve downstream reasoning \citep{ding2024entgpt}. Wang et al.\ compare LLM prompting strategies for entity matching and show that structured comparison and selection outperform naive generation, highlighting the importance of controlling the decision process \citep{wang2025match}. KG-assisted entity disambiguation similarly leverages existing graph structure to constrain candidate spaces and provide structured evidence \citep{pons2024knowledge}. At the extraction layer, schema-driven IE models such as GLiNER2 enable flexible extraction under user-specified label sets, demonstrating the practicality of schema-conditioned structured extraction \citep{zaratiana2025gliner2}. While these approaches improve grounding and extraction, they typically rely on external reference structures (KBs or predefined schemas) and are evaluated in isolation, rather than in end-to-end, corpus-scale settings where schema must be induced and entities and relations must be jointly canonicalized.

\paragraph{Context and conditions in knowledge representations.}
In scientific and technical domains, unconditional triples are often insufficient, as facts may hold only under specific conditions. Jiang et al.\ explicitly model \emph{conditions} to mitigate contradictions when aggregating scientific statements across incompatible contexts \citep{jiang2019role}. Context Graph extends this perspective by attaching structured contextual attributes (e.g., provenance, time, location, quantitative qualifiers) to relations and shows that treating context as first-class improves downstream reasoning \citep{xu2024context}. In applied domains such as mineral exploration, dynamic KGs similarly emphasize fusing heterogeneous evidence streams where interpretation depends on local contextual cues \citep{qin2025semantic}. Despite these advances, many construction pipelines still default to flat triples or weak qualifier structures, limiting reasoning about \emph{when} and \emph{under what assumptions} a relation holds. Wikontic~\cite{chepurova2026wikontic} further incorporates qualifiers into ontology-aware KG construction using Wikidata-aligned constraints. TRACE-KG builds on this broader direction but addresses a different setting in which contextual qualifiers are preserved while the entity and relation schemas are induced from the corpus rather than supplied by a predefined ontology.

\paragraph{Reliability, verification, provenance, and multimodal evidence.}
LLM-based extraction is vulnerable to hallucinations and structural inconsistencies, motivating constrained workflows and explicit verification stages. KARMA exemplifies this direction by coordinating specialized LLM agents for extraction, schema alignment, and conflict resolution when extending an existing KG \citep{lu2025karma}. Benchmarks such as \textsc{Text2KGBench} expose faithfulness and conformance failures even under schema constraints \citep{mihindukulasooriya2023text2kgbench}, while KGGen emphasizes retention and reuse under consolidation \citep{mo2025kggen}. However, many pipelines do not preserve fine-grained provenance as a first-class artifact—i.e., explicit links from each node or edge to supporting spans and structured elements such as tables, figures, and equations—and explicit confidence signals for prioritized, ad hoc review remain poorly standardized. These limitations are particularly pronounced in technical corpora, where critical information is distributed across narrative text and semi-structured or diagrammatic content.

Collectively, prior work advances key components of text-to-KG construction—ontology alignment, vocabulary consolidation, schema induction, entity matching, contextual modeling, and reliability-oriented orchestration. However, these components are largely developed in isolation and under incompatible assumptions about schema commitment, grounding, and evidence structure. The central gap is the open-world setting in which the schema itself must be induced from the corpus while entity identity, relation normalization, conditional context, and provenance remain jointly consistent across the entire construction process. TRACE-KG addresses this setting by integrating schema induction, resolution, contextualization, and traceability within a unified pipeline.

\section{Extended Methodology Details}
\label{app:method_details}

This appendix provides implementation-level details supporting reproducibility and auditing that are omitted from the main paper, including multi-field representation construction, clustering and subclustering strategies, constrained action interfaces (``function calling''), and iteration and stopping criteria. It also describes qualifier-aware relation merging and the traceability artifacts produced by the pipeline.

\subsection{Intermediate artifacts and traceability}
TRACE-KG persists intermediate JSONL artifacts at each stage to preserve traceability and enable inspection: chunk records (id, text, provenance); mention-level entity extractions; resolved entities with aggregated evidence and intrinsic properties; candidate and resolved entity classes with class groups; raw relation instances with qualifiers and evidence; and resolved relations with canonical labels and schema annotations. Each resolution stage additionally stores the exact prompt provided to the LLM, the raw LLM output for each processed cluster, and a structured action log.

\subsection{Multi-field representations and embedding}
Each resolution layer embeds \emph{multi-field representations} rather than single strings. Let $\phi(\cdot)$ denote the embedding function introduced in the preliminaries. Given a record with fields $\mathbf{f}=(f_1,\dots,f_k)$, we compute per-field embeddings $\phi(f_i)$ and combine them via a weighted sum:
\[
\Phi(\mathbf{f}) = \mathrm{norm}\!\left(\sum_{i=1}^k w_i \, \phi(f_i)\right),
\]
where $w_i$ are layer-specific hyperparameters.

\paragraph{Entity representations (EntRes).}
Fields include mention name, description, type hint, intrinsic-property strings, and short evidence or context snippets.

\paragraph{Entity class representations (EntClsRec/EntClsRes).}
Fields include class label, class description, evidence summaries, and member-entity summaries.

\paragraph{Relation representations (RelRes).}
Fields include raw relation label, relation description, endpoint context (subject and object names together with entity-schema metadata), coarse relation-type hints, and qualifiers.

\subsection{Clustering and subclustering}
TRACE-KG uses density-based clustering (HDBSCAN in our implementation) over representation vectors to form semantic neighborhoods. To keep LLM calls tractable, we apply two controls: (i) \textbf{local subclustering} for oversized clusters, where only members of a large cluster are reclustered; and (ii) \textbf{bounded prompt batching}, which processes at most $K$ items per LLM call by splitting a cluster into batches of size $K$.

\paragraph{Practical note.}
Clustering output is treated as \emph{suggestive}, not authoritative: the LLM may decide to merge none of the candidates, and multi-run reclustering can correct imperfect neighborhoods.

\subsection{Constrained action interfaces (``function calling'')}
In each resolution stage, the LLM outputs \emph{only} a JSON array of action objects selected from a fixed vocabulary. Deterministic code validates identifiers, executes edits, and logs every applied or rejected action.

\subsubsection{Entity resolution actions (EntRes)}
\begin{itemize}
  \item \textsc{MergeEntities}: merge multiple entity mentions into one canonical entity.
  \item \textsc{ModifyEntity}: revise the name, description, or type hint to prevent incorrect merges.
  \item \textsc{KeepEntity} (optional): explicitly keep a candidate unchanged.
\end{itemize}

\noindent\textbf{Illustrative action schemas.}
\begin{verbatim}
{ "action": "MergeEntities",
  "entity_ids": ["En_...","En_..."],
  "canonical_name": "...",
  "canonical_description": "...",
  "canonical_type": "...",
  "rationale": "..." }

{ "action": "ModifyEntity",
  "entity_id": "En_...",
  "new_name": "... or null",
  "new_description": "... or null",
  "new_type_hint": "... or null",
  "rationale": "..." }

{ "action": "KeepEntity",
  "entity_id": "En_...",
  "rationale": "..." }
\end{verbatim}

\subsubsection{Entity class resolution actions (EntClsRes)}
\begin{itemize}
  \item \textsc{merge\_classes}: merge redundant or synonymous classes.
  \item \textsc{split\_class}: split overloaded classes into coherent subclasses.
  \item \textsc{create\_class}: create a missing class for a coherent subset of entities.
  \item \textsc{reassign\_entities}: move entities between classes.
  \item \textsc{modify\_class}: revise class metadata, including class-group assignment.
\end{itemize}

Each action must include a one-line justification; optional confidence scores and remarks may also be provided. Newly created or merged classes may be referenced within the same output via provisional identifiers, which are resolved deterministically by the pipeline.

\subsubsection{Relation resolution actions (RelRes)}
\begin{itemize}
  \item \textsc{set\_canonical\_rel}: assign a canonical predicate label and description.
  \item \textsc{set\_rel\_cls}: assign a relation class.
  \item \textsc{set\_rel\_cls\_group}: assign a relation class group.
  \item \textsc{modify\_rel\_schema}: revise canonical and schema fields jointly.
  \item \textsc{add\_rel\_remark}: attach explanatory remarks without changing schema.
  \item \textsc{merge\_relations}: merge duplicate relation instances between identical endpoints (including direction-normalized cases), only when semantics are equivalent.
\end{itemize}

\paragraph{Macro relation groups.}
During RelRec, each relation instance includes a required single-token coarse group hint (\texttt{rel\_hint\_type}) selected from:
\{\texttt{IDENTITY}, \texttt{COMPOSITION}, \texttt{CAUSALITY}, \texttt{TEMPORALITY}, \texttt{SPATIALITY}, \texttt{ROLE}, \texttt{PURPOSE}, \texttt{DEPENDENCY}, \texttt{COUPLING}, \texttt{TRANSFORMATION}, \texttt{COMPARISON}, \texttt{INFORMATION}, \texttt{ASSOCIATION}\}. 
RelRes may revise this hint by assigning \texttt{rel\_cls\_group}.

\subsection{Qualifier extraction and normalization}
RelRec emits a qualifier dictionary with exactly eight fields:
\begin{verbatim}
{ "TemporalQualifier": ...,
  "SpatialQualifier": ...,
  "OperationalConstraint": ...,
  "ConditionExpression": ...,
  "UncertaintyQualifier": ...,
  "CausalHint": ...,
  "LogicalMarker": ...,
  "OtherQualifier": ... }
\end{verbatim}
Missing qualifiers are normalized to JSON \texttt{null} for consistency across relation instances.

\subsection{Safe duplicate relation merging and conflict handling}
When two relation instances are candidates for merging—i.e., they connect the same endpoints (after direction normalization) and express equivalent canonical predicates—TRACE-KG merges provenance and evidence while reconciling qualifiers conservatively.

\paragraph{Non-conflicting qualifiers.}
If one qualifier dictionary is a subset of the other, the merged relation retains the superset.

\paragraph{Conflicting qualifiers.}
If overlapping qualifier keys have different values (e.g., distinct temporal windows), both relation instances are retained, with an explanatory remark indicating the conflict.

\paragraph{Direction normalization.}
If duplicates are expressed in opposite directions, RelRes may normalize them to a single direction by swapping endpoints, while preserving evidence spans.

% \subsection{Iteration and stopping criteria}
% TRACE-KG applies iterative refinement in three stages.

% \paragraph{EntRes iteration.}
% Repeat: cluster $\rightarrow$ resolve $\rightarrow$ collapse until merges fall below a threshold or a maximum number of rounds is reached.

% \paragraph{EntClsRes multi-run refinement.}
% Repeat: cluster candidate classes $\rightarrow$ apply structural actions until structural edits plateau over a patience window or a maximum number of runs is reached.

% \paragraph{RelRes multi-run refinement.}
% Repeat: cluster relation instances $\rightarrow$ apply schema actions until schema edits plateau over a patience window or a maximum number of runs is reached.

\subsection{Iteration and stopping criteria}

The resolution stages use stage-specific stopping rules rather than a shared
``edit plateau'' criterion.

\paragraph{EntRes.}

Entity resolution repeats clustering, resolution, and representation refresh
until a round produces zero accepted \textsc{MergeEntities} actions or a
maximum of five iterations is reached.

\paragraph{EntClsRes.}

Entity-class resolution stops after three consecutive runs with zero accepted
structural edits, with a maximum of six runs.

\paragraph{RelRes.}

Relation resolution currently performs a single pass over the relation
clusters. Proposed relation canonicalization and schema actions are still
validated before deterministic execution.

\section{Detailed Experiment Descriptions}
\label{app:Ext_Experiment}

% ============================================================
% Appendix: Extended Experiment 1
% ============================================================
\subsection{Extended Experiment 1 Details (MINE-1)}
\label{sec:appendix_mine1}

\paragraph{Metric definitions.}
In addition to the standard MINE-1 retrieval accuracy (Ret.Acc), we evaluate
structural and representational properties of the generated graphs.

Ret.Acc remains the primary task metric; the remaining measures are reported
as complementary diagnostics of graph structure and representation.

\textbf{Connectivity (Conn.)} is the fraction of nodes in the largest weakly connected component \cite{newman2003structure}. 
\textbf{Average degree (AvgDeg)} is $|E|/|V|$ \cite{newman2010networks}. 
\textbf{Clustering coefficient (Clust.)} measures average local clustering \cite{watts1998collective,newman2010networks}. 
\textbf{Leakage (Leak\%)} quantifies lexical overlap between source text and entity strings using 4-gram overlap \cite{lin2004rouge}. 
\textbf{Triple compression ratio (TriCR)} is the ratio of total triple word count to source word count; values near 1 indicate balanced compression, while deviations indicate information loss or redundancy \cite{shannon1948mathematical}.

\paragraph{Composite metrics.}
RWA (Reachability-Weighted Accuracy) adjusts retrieval accuracy by connectivity,
reflecting that disconnected components cannot support multi-hop retrieval.
EGU (Effective Graph Utilization) further discounts retrieval by leakage,
reducing credit for verbatim copying \cite{van1979information}.
SCI (Structural Coherence Index) measures structural quality independently of
retrieval accuracy by combining relational density, clustering, and connectivity.

SCI is intended as a relative conjunctive diagnostic rather than an absolute
measure of graph quality. Its multiplicative form requires all three structural
properties to contribute jointly: if relational density, local clustering, or
global connectivity is near zero, the resulting SCI decreases accordingly.
We therefore interpret SCI comparatively across graphs and report its
components alongside it rather than treating the scalar value in isolation.

\textbf{Average rank (AvgRank)} measures the average rank position of the first retrieved supporting evidence for each query; lower values indicate more efficient retrieval.

\paragraph{Structural diagnostics.}
Table~\ref{tab:mine1_struct} complements Table~\ref{tab:mine1_main} by exposing graph properties underlying retrieval performance: graph size ($|V|$, $|E|$), entity granularity (AvgEW), relational density (AvgDeg), global connectivity (Conn.), and local cohesion (Clust.). While these are descriptive rather than direct objectives, they explain why methods with similar Ret.Acc diverge after accounting for leakage and reachability.

TRACE-KG combines compact entities (AvgEW = 2.4) with the highest average degree and clustering, indicating dense and locally coherent neighborhoods. This aligns with its strong SCI and EGU scores in the main table. GraphRAG achieves the highest connectivity, but on much smaller graphs, suggesting connectivity driven by compression rather than broad relational coverage. AutoSchemaKG produces substantially longer entities (AvgEW = 6.6), consistent with its high leakage. OpenIE yields the largest graphs ($|V|$, $|E|$), but low clustering indicates weak local structure despite expansion. KGGen produces smaller, sparse graphs, explaining its low connectivity and reachability-adjusted performance.

\begin{table*}[t]
\caption{Additional structural diagnostics complementary to Table~\ref{tab:mine1_main}. $|V|$ and $|A|$ denote average node and edge counts; AvgEW is average words per entity. AvgDeg is $|E|/|V|$ \cite{newman2010networks}; Conn.\ is largest weakly connected component fraction \cite{newman2003structure}; Clust.\ is average clustering coefficient \cite{watts1998collective,newman2010networks}. Higher is better only for AvgDeg, Conn., and Clust.}
\centering
\small
\setlength{\tabcolsep}{6pt}
\begin{tabular}{lcccccc}
\toprule
Method & $|V|$ & $|E|$ & AvgEW & AvgDeg$\uparrow$ & Conn.$\uparrow$ & Clust.$\uparrow$ \\
\midrule
TRACE-KG     & 63  & 90  & 2.4 & \textbf{1.35} & 88.5\%          & \textbf{0.199} \\
GraphRAG     & 11  & 11  & 1.6 & 0.98          & \textbf{91.5\%} & 0.150 \\
AutoSchemaKG & 107 & 105 & 6.6 & 0.98          & 61.5\%          & 0.066 \\
OpenIE       & 170 & 253 & 3.2 & 1.26          & 74.0\%          & 0.027 \\
KGGen        & 84  & 65  & 1.2 & 0.75          & 46.1\%          & 0.011 \\
\bottomrule
\end{tabular}
\label{tab:mine1_struct}
\end{table*}

\paragraph{Baseline behavior and trade-offs.}
AutoSchemaKG achieves the highest Ret.Acc but relies on lexical retention, reflected in high leakage and long entity strings. TRACE-KG instead maintains compact entities and strong structural properties, achieving the highest AvgDeg and clustering.

GraphRAG achieves high connectivity but on small graphs ($|V|=11$, $|E|=11$), limiting coverage and reducing Ret.Acc. OpenIE expands the source heavily (TriCR = 3.201) with weak structure, whereas KGGen compresses aggressively (TriCR = 0.494) but produces fragmented graphs (Conn. = 46.1\%). These patterns highlight trade-offs between lexical retention, compression, and structural coherence.

\begin{figure}[t]
\centering
\includegraphics[width=0.92\linewidth]{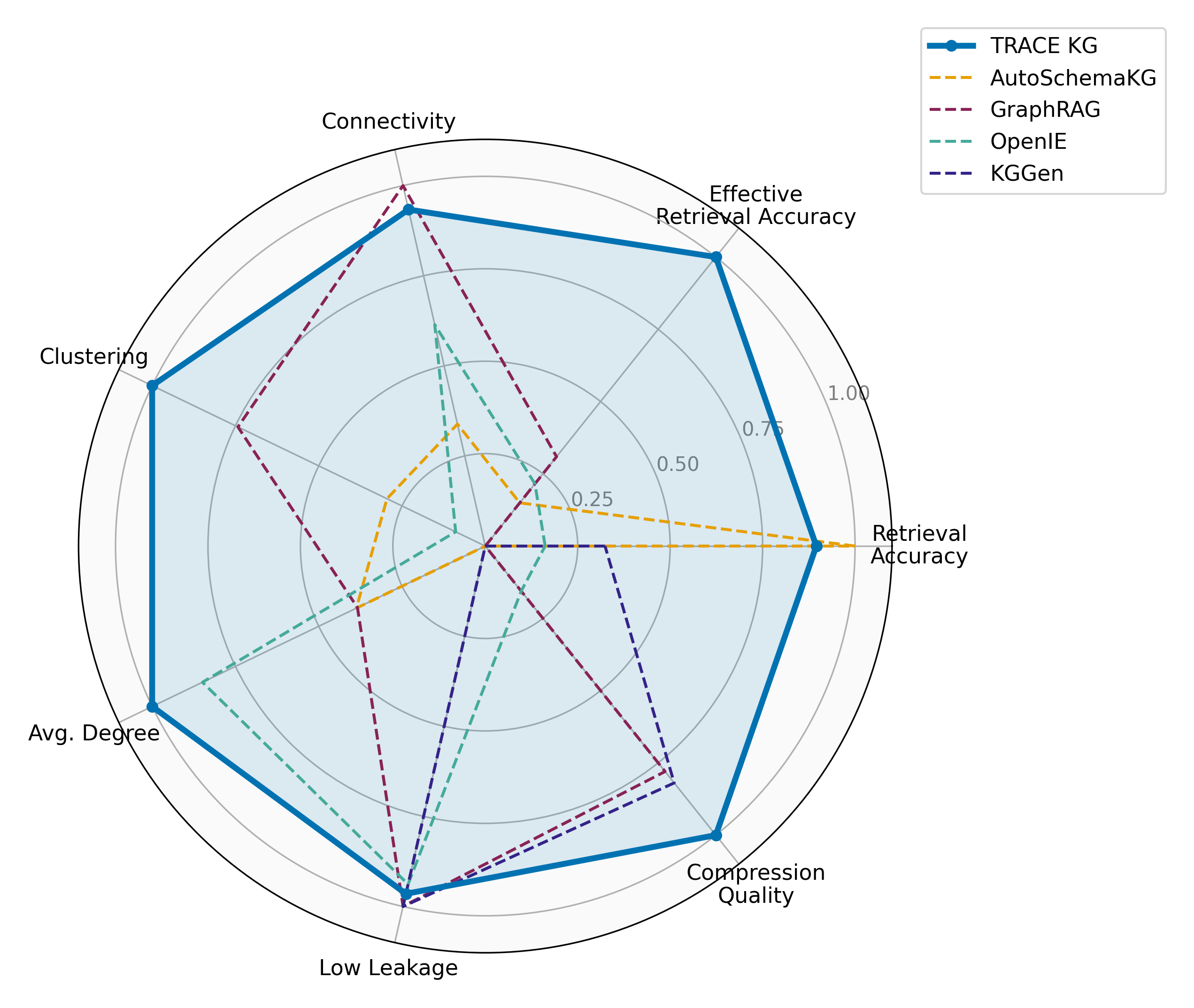}
\vspace{-4pt}
\caption{Normalized multi-metric profile on MINE-1. Leak is inverted, and compression is expressed as proximity to TriCR = 1 so that higher values are preferable on all axes.}
\label{fig:mine1_radar}
\vspace{-8pt}
\end{figure}

\paragraph{Supplementary visualizations.}
Figure~\ref{fig:mine1_radar} provides a normalized multi-metric view. It is diagnostic rather than primary and should be interpreted alongside Table~\ref{tab:mine1_main}. The radar view shows that TRACE-KG remains consistently strong across retrieval accuracy, effective retrieval after discounting, connectivity, clustering, and compression quality, rather than excelling on a single axis.

\begin{figure}[t]
\centering
\includegraphics[width=0.88\linewidth]{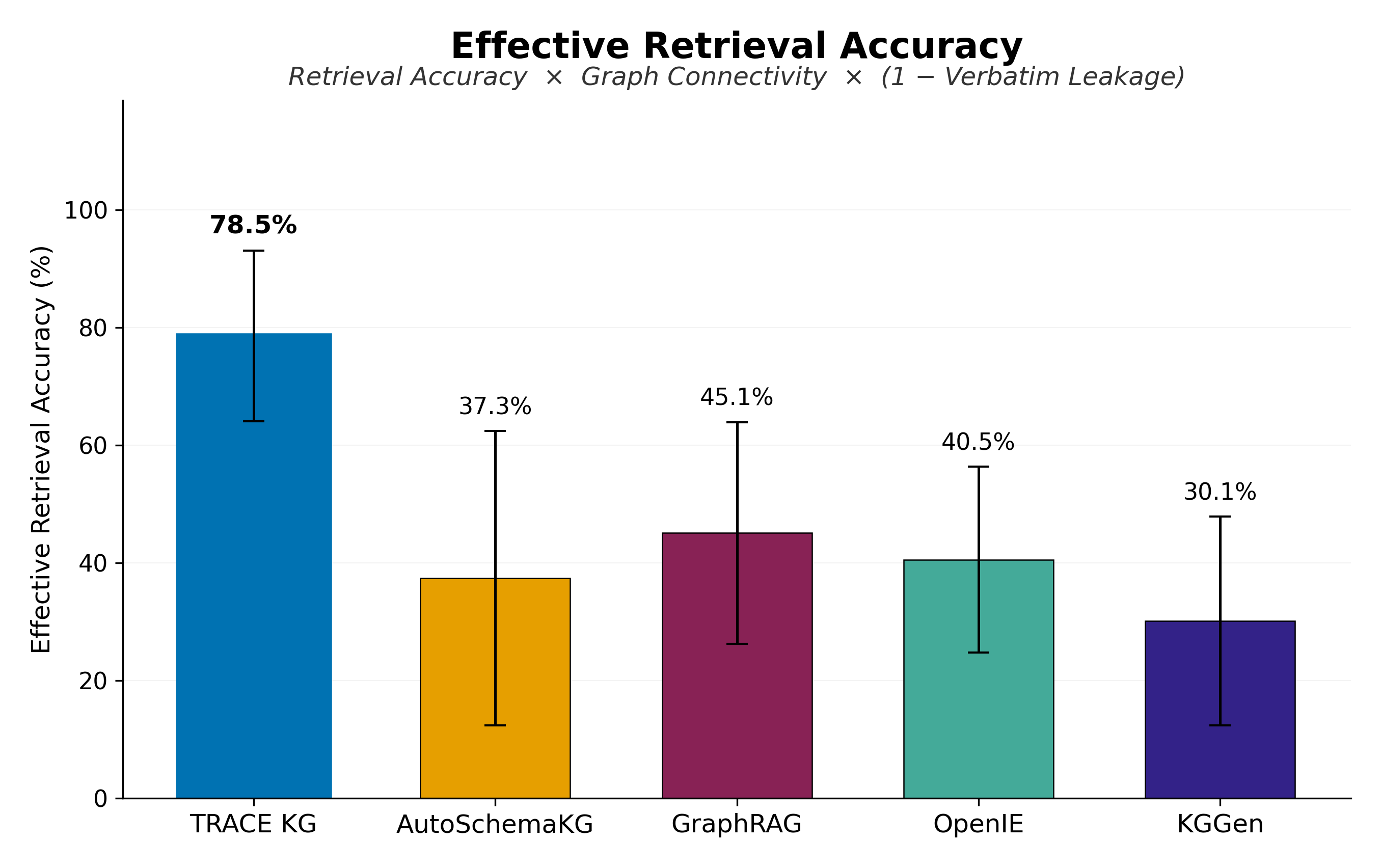}
\vspace{-4pt}
\caption{Effective Graph Utilization (EGU) on MINE-1. Error bars indicate variability across benchmark instances.}
\label{fig:mine1_egu_variation}
\vspace{-8pt}
\end{figure}

Figure~\ref{fig:mine1_egu_variation} visualizes EGU variability across benchmark instances. Its purpose is to show that the observed performance gap is stable and not driven by outliers; the ordering in Table~\ref{tab:mine1_main} remains consistent across instances.

% =========================================================
% Experiment 2 (Appendix / Extended)
% =========================================================
\subsection{Extended Experiment 2 Details}
\label{app:exp2_schema_details}

\paragraph{Dataset and evaluation setup.}
We use the DBpedia-WebNLG collection distributed in Text2KGBench/OSKGC \cite{mihindukulasooriya2023text2kgbench, wang2025oskgc}, which provides $19$ domain ontologies paired with sentence-level gold triples (4,860 sentences in total). Table~\ref{tab:exp2_dataset_stats} summarizes per-domain statistics. 

For each domain, TRACE-KG is run once on the \texttt{train} sentences to induce a fixed TRACE schema. Schema-to-ontology alignment judgements are then computed once per domain and reused across evaluation scopes. The three scopes in the main paper (\textbf{Source}, \textbf{Held-out}, \textbf{Combined}) differ only in which gold triples activate reference anchors and in their frequency weights; the induced TRACE schema and alignment mapping remain fixed.

\paragraph{Schema mapping.}
Schema mapping is performed at the schema level after induction. Each reference ontology anchor is aligned against the induced TRACE schema rather than sentence-level extractions. Because TRACE induces hierarchical schema structures, valid correspondences may occur at multiple abstraction levels. We therefore preserve the hierarchy during evaluation instead of flattening the schema into a single label space.

\paragraph{Candidate retrieval.}
For each reference anchor, we retrieve a small candidate set from the induced TRACE schema prior to semantic verification. Retrieval uses cosine similarity over weighted multi-evidence embeddings. 

Entity-side evidence includes schema labels across hierarchy levels and representative instance cues. Relation-side evidence includes canonical labels, higher-level relation classes, lexical variants, and representative subject--object examples with induced type signatures. Reference-side evidence includes ontology labels, domain/range constraints, and examples from \texttt{train} triples. 

We retrieve $K{=}5$ candidates per anchor and apply a controlled assignment step to limit fan-out and reduce spurious many-to-many correspondences. For reproducibility, we persist the induced schema, candidate lists with similarity scores, LLM prompts and outputs, parsed judgements, and per-domain summaries.

\paragraph{LLM-based alignment and audit.}
Lexical similarity alone is insufficient for schema alignment under synonymy and granularity variation. We therefore use an LLM as a structured verifier for all retrieved (reference anchor, TRACE candidate) pairs. Each candidate is presented together with its hierarchical context, allowing alignment at the most appropriate abstraction level.

The judge assigns one of \textsc{Equivalent}, \textsc{Narrower}, \textsc{Broader}, or \textsc{Unrelated}, along with a confidence score. \textsc{Equivalent} is treated as exact recovery and \textsc{Narrower} as compatible refinement; \textsc{Broader} and \textsc{Unrelated} do not contribute to coverage. For relations, alignment is evaluated in a direction-relaxed manner by considering both orientations and retaining the better match.

The verifier is run deterministically (temperature $0$). Low-confidence cases (confidence $<0.88$) undergo targeted human audit to ensure that aggregate trends are not artifacts of unstable judgements.

\paragraph{Active anchors and metrics.}
Metrics are computed under three scopes: \textbf{Source}, \textbf{Held-out}, and \textbf{Combined}. In each scope, only \emph{active} reference anchors are evaluated: ontology relations appearing in the scope's gold triples together with their domain and range concepts. Anchors are frequency-weighted according to scope-specific gold triple counts \cite{euzenat2005towards}.

We report \textbf{Coverage} (Exact and Compatible), frequency-weighted \textbf{MRR@5}, and \textbf{Domain/Range Consistency} (D/R). The D/R check is direction-relaxed and hierarchy-aware, allowing backoff to coarser TRACE types when fine-grained typing is sparse. Primitive datatypes (e.g., \texttt{xsd:date}, \texttt{xsd:string}) are excluded, as they cannot be induced as text-derived classes.

\begin{table}[t]
\centering
\caption{DBpedia-WebNLG statistics for Experiment~2.}
\label{tab:exp2_dataset_stats}
\small
\setlength{\tabcolsep}{6pt}
\begin{tabular}{r l r r r}
\toprule
\textbf{\#} & \textbf{Ontology} & \textbf{\#Types} & \textbf{\#Relations} & \textbf{\#Sentences} \\
\midrule
1  & University     & 15 & 46 & 156 \\
2  & Music          & 15 & 35 & 290 \\
3  & Airport        & 14 & 39 & 306 \\
4  & Building       & 14 & 38 & 275 \\
5  & Athlete        & 17 & 37 & 293 \\
6  & Politician     & 19 & 40 & 319 \\
7  & Company        & 10 & 28 & 153 \\
8  & Celestial      &  8 & 27 & 194 \\
9  & Astronaut      & 16 & 38 & 154 \\
10 & Comics         & 10 & 18 & 102 \\
11 & Transport      & 20 & 68 & 314 \\
12 & Monument       & 14 & 26 &  92 \\
13 & Food           & 12 & 24 & 398 \\
14 & Written Work   & 10 & 44 & 322 \\
15 & Sports Team    & 14 & 24 & 235 \\
16 & City           & 11 & 23 & 348 \\
17 & Artist         & 20 & 39 & 386 \\
18 & Scientist      & 15 & 47 & 259 \\
19 & Film           & 18 & 44 & 264 \\
\midrule
\multicolumn{4}{r}{\textbf{Total Sentences}} & \textbf{4,860} \\
\bottomrule
\end{tabular}
\end{table}

\paragraph{Granularity and alignment behavior.}
Because TRACE induces a hierarchy rather than a flat label space, successful mappings may occur at different abstraction levels. Table~\ref{tab:exp2_granularity} shows that most relation alignments occur at the finest level, while concept alignments are more evenly distributed across levels. This supports the interpretation in the main paper that many successful matches arise through compatible refinement rather than exact equivalence.

\begin{table}[t]
\centering
\caption{Best-matching TRACE level among accepted alignments (macro-averaged, Combined).}
\label{tab:exp2_granularity}
\small
\setlength{\tabcolsep}{6pt}
\begin{tabular}{lcc}
\toprule
\textbf{Level} & \textbf{Relations} & \textbf{Concepts} \\
\midrule
Finest level   & 64.2\% & 50.8\% \\
Middle level   & 28.7\% & 35.4\% \\
Coarsest level &  7.1\% & 13.8\% \\
\bottomrule
\end{tabular}
\end{table}

\paragraph{Interpretation of alignment results.}
The Exact--Compatible gap in Table~\ref{tab:exp2_main} primarily reflects granularity mismatch rather than alignment failure. TRACE often induces more specific schema elements than the reference ontology, leading to \textsc{Narrower} matches. Thus, compatible coverage is a more appropriate indicator of successful schema recovery than exact equivalence alone.

A second pattern is that retrieval is not the primary bottleneck: MRR@5 remains high, indicating that compatible candidates are typically retrieved near the top. The larger drop in relation performance under the Held-out setting reflects the difficulty of stabilizing relation canonicalization under sparse evidence and lexical variability.

\paragraph{Failure modes and structural consistency.}
Residual failures concentrate in two regimes: (i) low-frequency relations with insufficient evidence to stabilize canonicalization, and (ii) semantically adjacent predicates whose distinction depends on context not fully captured at the sentence level. In both cases, \textsc{Narrower} matches dominate, reflecting consistent but more fine-grained schema induction.

D/R consistency is computed only over relations with compatible mappings and therefore reflects structural coherence rather than coverage. The high D/R values indicate that, once aligned, TRACE relations connect semantically appropriate endpoint types, supporting the claim that the induced schema is structurally consistent and reusable.

\subsection{Cross-Document Consolidation Details}
\label{app:crossdoc}

\begin{table}[t]
\centering
\footnotesize
% \color{blue}
\caption{Cross-document consolidation on the ISO~55000-family corpus.
Processing the three standards separately (summing outputs) versus jointly as
one corpus. Reduction reflects vocabulary compression from cross-document
consolidation.}
\label{tab:iso_cross_document_detail}
\setlength{\tabcolsep}{5pt}
\renewcommand{\arraystretch}{1.05}
\begin{tabular}{lccc}
\toprule
\textbf{Metric} & \textbf{Separate} & \textbf{Combined} & \textbf{Reduction} \\
\midrule
Pages / chunks          & 83 / 339 & 83 / 339 & -- \\
Entity mentions         & 638      & 638      & -- \\
Raw relation candidates & 882      & 847      & 4.0\% \\
\midrule
Resolved entities       & 403      & 285      & 29.3\% \\
Entity classes          & 124      & 64       & 48.4\% \\
Canonical relations     & 229      & 156      & 31.9\% \\
Relation classes        & 113      & 62       & 45.1\% \\
\bottomrule
\end{tabular}
\end{table}

\begin{table}[t]
\centering
\footnotesize
% \color{blue}
\caption{Entity resolution audit on the ISO~55000-family corpus (manual
inspection). Precision is measured over audited resolution actions; recall is
measured over sampled resolved entities.}
\label{tab:iso_resolution_audit}
\setlength{\tabcolsep}{5pt}
\renewcommand{\arraystretch}{1.05}
\begin{tabular}{llcc}
\toprule
\textbf{Axis} & \textbf{Outcome} & \textbf{Count} & \textbf{Rate} \\
\midrule
\multirow{3}{*}{Precision}
 & Correct (valid merge / keep / modify) & 222/249 & 89.2\% \\
 & False merge (distinct entities merged) & 11/249 & 4.4\% \\
 & Incorrect modify (attribute distortion) & 16/249 & 6.4\% \\
\midrule
Recall
 & Missed duplicates (should have merged) & 9/120 & 7.5\% \\
\bottomrule
\end{tabular}
\end{table}

\paragraph{Setting.}
Experiment~2 uses DBpedia-WebNLG/Text2KGBench because its held-out
human-designed ontologies across diverse domains enable systematic schema
evaluation, but its sentence-level inputs do not test cross-document
consolidation. We therefore evaluate TRACE-KG on three related technical
standards from the ISO 55000 asset-management family (ISO 55000, ISO 55001,
and ISO 55002; 83 pages, 339 chunks) that share terminology across documents.

\paragraph{Protocol.}
We compare processing the three standards independently and summing their
outputs against processing them jointly as one corpus. Document inputs,
chunking, and entity mention extraction are unchanged; relation candidates may
differ because relation recognition is conditioned on resolved entities. As
shown in Table~\ref{tab:iso_cross_document_detail}, joint processing yields a
substantially more compact vocabulary---reducing resolved entities by 29.3\%,
entity classes by 48.4\%, canonical relations by 31.9\%, and relation classes
by 45.1\%---despite identical input pages, chunks, and mentions. This reduction
reflects TRACE-KG recognizing and consolidating terminology shared across the
three standards rather than treating each document in isolation.

\paragraph{Entity resolution audit.}
To verify that this reduction reflects valid consolidation rather than
indiscriminate merging, and to assess resolution correctness directly rather
than only through downstream retrieval, we manually audited resolution
decisions along two axes (Table~\ref{tab:iso_resolution_audit}). For precision,
we audited 249 resolution actions: 222 (89.2\%) were correct, with only 11
false merges (4.4\%) and 16 incorrect modifications (6.4\%). For recall, a
separate audit of 120 sampled resolved entities found 9 residual duplicates
(7.5\%). The low false-merge rate indicates precision-oriented consolidation
that favors valid merges over indiscriminate collapsing, consistent with the
constrained, neighborhood-bounded resolution design.

\subsection{Ablation and Backbone Sensitivity Analysis}
\label{app:ablation_analysis}

\begin{table}[t]
\centering
\caption{Component ablation and backbone sensitivity on 25 MINE-1
articles (375 fact checks).}
\label{tab:ablation}
\footnotesize
\setlength{\tabcolsep}{5pt}
\renewcommand{\arraystretch}{0.95}

\begin{tabular}{lc}
\toprule
\textbf{Setting} & \textbf{Ret.Acc} \\
\midrule

\multicolumn{2}{l}{\textit{Component ablation}} \\
Full TRACE-KG                 & 91.4\% \\
w/o qualifier extraction      & 82.4\% \\
w/o iterative entity res.     & 84.3\% \\
w/o relation canonicalization & 86.0\% \\
w/o schema induction          & 88.2\% \\

\midrule
\multicolumn{2}{l}{\textit{Backbone sensitivity}} \\
Gemini 2.0 Flash  & 75.1\% \\
GPT-4o            & 83.2\% \\
Claude Sonnet 3.5 & 86.7\% \\
GPT-5             & 91.4\% \\

\bottomrule
\end{tabular}
\end{table}

\paragraph{Component ablation.}
Table~\ref{tab:ablation} reports component ablations on 25 of the 100 MINE-1
articles (375 fact checks), using identical retrieval parameters, embeddings,
and judge across settings. This subset is slightly easier than the full
benchmark, yielding a full-model Ret.Acc of 91.4\% compared with 90.2\% on
the complete benchmark. Removing qualifier extraction causes the largest
Ret.Acc drop, suggesting that qualifiers materially affect whether retrieved
evidence preserves the context needed to support a fact. Entity resolution
and relation canonicalization also contribute substantially by reducing entity
fragmentation and predicate inconsistency. Schema induction has a smaller
effect on MINE-1, consistent with the benchmark's emphasis on local factual
support rather than corpus-level organization.

\paragraph{Backbone sensitivity.}
Table~\ref{tab:ablation} also reports backbone sensitivity on the same 25
MINE-1 articles (375 fact checks), keeping the pipeline, retrieval parameters,
embeddings, and judge fixed across settings. Performance improves substantially
with stronger backbones, indicating that TRACE-KG is sensitive to the
capability of the underlying LLM. At the same time, the pipeline operates
across all four model families, showing that its construction procedure is not
specific to GPT-5.

\subsection{Faithfulness and Hallucination Audit}
\label{app:faithfulness_audit}

\begin{table}[t]
% \color{blue}
\centering
\footnotesize
\caption{Faithfulness audit on 25 MINE-1 articles (GPT-5 backbone), by manual
inspection. We sample 64 relation instances---32 from successful and 32 from
failed fact checks---and judge each triplet element (head, relation, tail)
against its cited provenance chunk.}
\label{tab:faithfulness_audit}
\setlength{\tabcolsep}{4pt}
\renewcommand{\arraystretch}{1.05}
\begin{tabular}{llcccc}
\toprule
\textbf{Group} & \textbf{Element} & \textbf{Grounded} & \textbf{Over-gen.} & \textbf{Distorted} & \textbf{Halluc.} \\
\midrule
\multirow{4}{*}{Success}
 & Head      & 30/32 & 2/32 & 0/32 & 0/32 \\
 & Relation  & 28/32 & 2/32 & 2/32 & 0/32 \\
 & Tail      & 30/32 & 2/32 & 0/32 & 0/32 \\
 & \textit{Subtotal} & \textit{88/96} & \textit{6/96} & \textit{2/96} & \textit{0/96} \\
\midrule
\multirow{4}{*}{Failure}
 & Head      & 27/32 & 3/32 & 1/32 & 1/32 \\
 & Relation  & 23/32 & 3/32 & 4/32 & 2/32 \\
 & Tail      & 27/32 & 3/32 & 1/32 & 1/32 \\
 & \textit{Subtotal} & \textit{77/96} & \textit{9/96} & \textit{6/96} & \textit{4/96} \\
\bottomrule
\end{tabular}
\end{table}

\paragraph{Setup.}
To directly assess whether extracted content is grounded in source evidence,
and to characterize the role of hallucination, we conducted a manual
faithfulness audit on the 25-article MINE-1 subset (GPT-5 backbone; 375 fact
checks). To avoid introducing an additional LLM evaluation layer, all
judgements were made by manual inspection. We sampled 64 relation instances in
two groups: 32 from successful fact checks (randomly sampled) and 32 from
failed fact checks. For each relation instance we inspected all three elements
of the triplet---head entity, relation predicate, and tail entity---against the
chunk cited as its provenance, giving up to 96 element-level judgements per
group.

\paragraph{Judgement categories.}
Each element was assigned to exactly one of four mutually exclusive categories:
\textbf{Fully Grounded} (explicitly and accurately supported by the cited
chunk), \textbf{Over-generalized} (correct but broader or less specific than
the source), \textbf{Distorted} (present in the source but with altered
meaning, e.g., a shifted predicate sense or dropped condition), and
\textbf{Hallucinated} (no textual basis in the cited chunk). \emph{Distorted}
and \emph{Hallucinated} are distinct: the former reflects an inaccurate
rendering of content that exists, whereas the latter reflects fabricated
content.

\paragraph{Findings.}
Among successful fact checks, 91.7\% of elements were fully grounded with
\emph{no hallucinations}, and the few errors were confined to over-generalization
and mild predicate distortion. Among failures, grounding decreased (80.2\%) and
hallucination appeared but remained low (4.2\%), with distortion and
over-generalization accounting for a larger share of errors than fabrication.
These results indicate that hallucination is not the dominant failure mode:
even when a fact check fails, the underlying triplets are usually present in the
source but imperfectly rendered. This is consistent with TRACE-KG's constrained
design---chunk-local extraction, structured actions over existing identifiers,
deterministic validation, and provenance-grounded elements---which limits
fabrication and keeps residual errors inspectable against their supporting
evidence.

\section{Implementation Details}
\label{app_sec:implementation}

\paragraph{Setup.}
% An anonymized implementation of TRACE-KG is available at
% \url{https://github.com/mosadddeq01/TRACE-KG}, with additional code included in the supplementary material. 
The implementation of TRACE-KG is available at
\url{https://github.com/mosadddeq01/TRACE-KG}. All experiments were conducted on a single compute node with one GPU (\texttt{gpu:1g.20gb:1}), 16 CPU cores, and 40\,GiB RAM.

\paragraph{Pipeline configuration.}
TRACE-KG is implemented in Python and combines sentence-preserving chunking, transformer-based embeddings, density-based clustering, and step-specific LLM prompting within a unified pipeline. Unless otherwise specified, all stages use \texttt{gpt-5.1} with a maximum token budget of 16{,}000. Multimodal ingestion uses a vision-capable model (\texttt{gpt-5.1}) to convert non-text elements into structured text.

\paragraph{Chunking and context.}
Documents are segmented into sentence-preserving chunks of 100--200 tokens using spaCy (\texttt{en\_core\_web\_sm}). Entity recognition operates at the chunk level with a context window of up to four preceding chunks.

\paragraph{Embeddings and clustering.}
Multi-field representations are embedded using \texttt{BAAI/bge-large-en-v1.5} (batch size 32, mean pooling, L2 normalization). Semantic neighborhoods are formed with HDBSCAN, with optional UMAP preprocessing. LLM calls are bounded by batching at most 10 items per prompt.

\paragraph{LLM interaction.}
LLM decisions are constrained through structured action interfaces, while execution is deterministic and validated by the pipeline. Entity recognition uses an 8{,}000-token limit, while resolution stages use 16{,}000 tokens. Temperature is omitted for GPT-5 models to ensure stable outputs. We use six stage-specific prompt templates—\textit{Entity Recognition}, \textit{Entity Resolution}, \textit{Entity Class Recognition}, \textit{Entity Class Resolution}, \textit{Relation Recognition}, and \textit{Relation Resolution}—defined in \texttt{TKG\_Prompts.py}; all prompts are available at \url{https://github.com/mosadddeq01/TRACE-KG}.

\paragraph{Experiment-specific settings.}
For Experiment~1, retrieval uses top-$k=8$ entities with a 2-hop expansion (up to 250 nodes and 300 edges), and evaluation uses a fixed LLM judge (\texttt{gpt-5.1}). For Experiment~2, candidate retrieval uses top-$K=5$ schema elements per anchor; alignment uses an LLM verifier (1{,}400-token limit, top-$k=6$), with anchored matching threshold 0.20 and at most 3 assignments per element. Low-confidence outputs (confidence $<0.88$) are subject to targeted human audit.

\paragraph{Computational cost.}
Table~\ref{tab:cost} reports the end-to-end computational cost of TRACE-KG on
the combined ISO multi-document experiment (Section~\ref{sec:iso_cross_document};
0.61M source characters), covering the full construction pipeline: recognition,
schema induction, and resolution. To keep cost tractable as the corpus grows,
TRACE-KG applies two explicit controls (Appendix~\ref{app:method_details}):
local subclustering for oversized clusters and bounded prompt batching (at most
10 items per LLM call), which cap the number and size of LLM calls rather than
letting them scale with cluster size. We report both raw totals and figures
normalized per 1M source characters to support extrapolation to larger corpora.

\begin{table}[t]
\centering
\footnotesize
% \color{blue}
\caption{End-to-end computational cost of TRACE-KG on the combined ISO
multi-document experiment (full pipeline: recognition, schema induction,
resolution). Normalized figures are per 1M source characters.}
\label{tab:cost}
\setlength{\tabcolsep}{6pt}
\renewcommand{\arraystretch}{1.05}
\begin{tabular}{lcc}
\toprule
\textbf{Metric} & \textbf{Raw (0.61M chars)} & \textbf{Normalized (per 1M chars)} \\
\midrule
LLM calls       & 876   & 1{,}436 \\
Input tokens    & 3.60M & 5.90M   \\
Output tokens   & 0.42M & 0.69M   \\
Wall-clock time & 2.8 h & 4.6 h   \\
\bottomrule
\end{tabular}
\end{table}

\section{Limitations and Future Work}

TRACE-KG remains sensitive to semantic-neighborhood quality and LLM capability,
and the ISO experiment does not establish scalability to substantially larger
or more heterogeneous corpora. RWA, EGU, and SCI remain diagnostic metrics
whose relationship to downstream KG utility requires further validation.
Finally, multimodal ingestion relies on VLM-based textualization rather than
direct cross-modal reasoning, so description errors may propagate into KG
construction despite provenance-based inspection.

\section{Case Study}
\label{app:casestudy}

\paragraph{Document and motivation.}
% As a running case study, we use \emph{SEMI Document 6578}, a revision to \emph{SEMI E10-0814E}, which specifies the definition and measurement of equipment reliability, availability, maintainability (RAM), and utilization. The document is a strong stress test for TRACE-KG because it is both technically dense and multimodal: even its opening pages combine formal purpose/scope statements, mutually exclusive equipment states, subsystem and cluster-tool terminology, and metric definitions such as MTBF, MCBF, MWBF, uptime, maintainability, and utilization.
 As a running case study, we use \emph{SEMI Document 6578}, a revision to \emph{SEMI E10-0814E}, which specifies the definition and measurement of equipment reliability, availability, maintainability (RAM), and utilization. Maintenance and reliability records are a natural target for knowledge-graph construction because prior work has shown that such knowledge is difficult to reuse when it remains in noisy, semi-structured, or non-computable records rather than formalized semantic representations \cite{ameri6017381tabular}. The document is a strong stress test for TRACE-KG because it is both technically dense and multimodal: even its opening pages combine formal purpose/scope statements, mutually exclusive equipment states, subsystem and cluster-tool terminology, and metric definitions such as MTBF, MCBF, MWBF, uptime, maintainability, and utilization. More broadly, this case study reflects long-context document understanding settings in which evidence is distributed across text, layout, tables, charts, and images rather than concentrated in a single local passage \citep{ma2024mmlongbench}.

\paragraph{Why this document is challenging.}
% The document mixes several levels of abstraction that are difficult to organize with flat triple extraction alone. It refers to equipment systems, subsystems, modules, states, substates, downtime categories, and performance measures, while also specifying how these concepts interact. In addition, essential information is distributed across prose, equations, and diagrams rather than appearing in a single textual form. A useful representation therefore requires more than local extraction: it must consolidate repeated technical concepts, distinguish closely related but non-identical terms, and preserve provenance to the original evidence.
The document mixes several levels of abstraction that are difficult to organize with flat triple extraction alone. It refers to equipment systems, subsystems, modules, states, substates, downtime categories, and performance measures, while also specifying how these concepts interact. Such hierarchical dependencies, dense technical terminology, and limited availability of external reference knowledge are characteristic challenges in domain-specific KG construction \citep{sun2025lkd}. In addition, essential information is distributed across prose, equations, and diagrams rather than appearing in a single textual form. A useful representation therefore requires more than local extraction: it must consolidate repeated technical concepts, distinguish closely related but non-identical terms, preserve provenance to the original evidence, and make granularity choices explicit. Analogous granularity issues have been observed in LLM-based log parsing, where different levels of specificity and applicability can yield different but plausible structured outputs \citep{zhong2024logparser}.

\paragraph{Multimodal-ingestion ablation.}
We evaluate the contribution of non-text document evidence on a subset of the
case study by comparing PDF text alone with PDF text augmented by VLM-generated
descriptions of figures, tables, diagrams, and equations. Table~\ref{tab:mm_ablation}
summarizes the results. Non-text-grounded facts are those whose supporting
provenance originates from a non-text source region.
\begin{table}[t]
\centering
\caption{Effect of multimodal ingestion on the SEMI case-study subset.}
\label{tab:mm_ablation}
\small
\begin{tabular}{lrrr}
\toprule
\textbf{Setting} &
\textbf{Entities} &
\textbf{Relations} &
\textbf{Non-text facts} \\
\midrule
Text only        & 214 & 336 & 0  \\
Text + VLM       & 241 & 389 & 47 \\
\bottomrule
\end{tabular}
\end{table}
Adding VLM-generated descriptions increases resolved entities by 12.6\% and
grounded relations by 15.8\%, while recovering 47 facts grounded in non-text
regions. Manual inspection found 42/47 of these facts supported by their
source regions, indicating improved coverage while also motivating
provenance-based inspection of VLM-derived evidence.

\begin{figure*}[t]
    \centering
    \includegraphics[width=\textwidth]{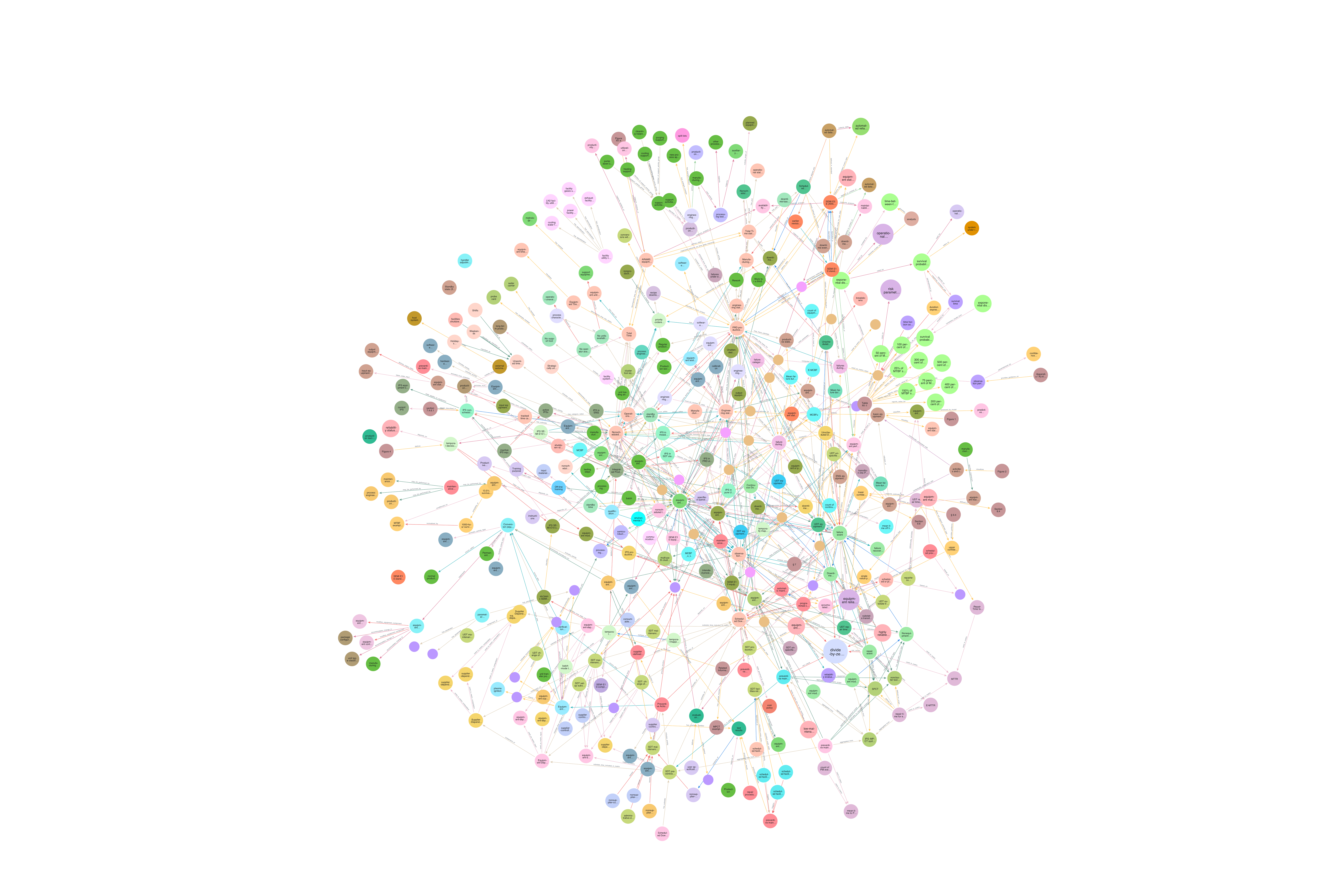}
    \caption{Overview of the constructed context-enriched knowledge graph for the case-study document.}
    \label{fig:case-wholekg}
\end{figure*}

\begin{figure*}[t]
    \centering
    \includegraphics[width=\textwidth]{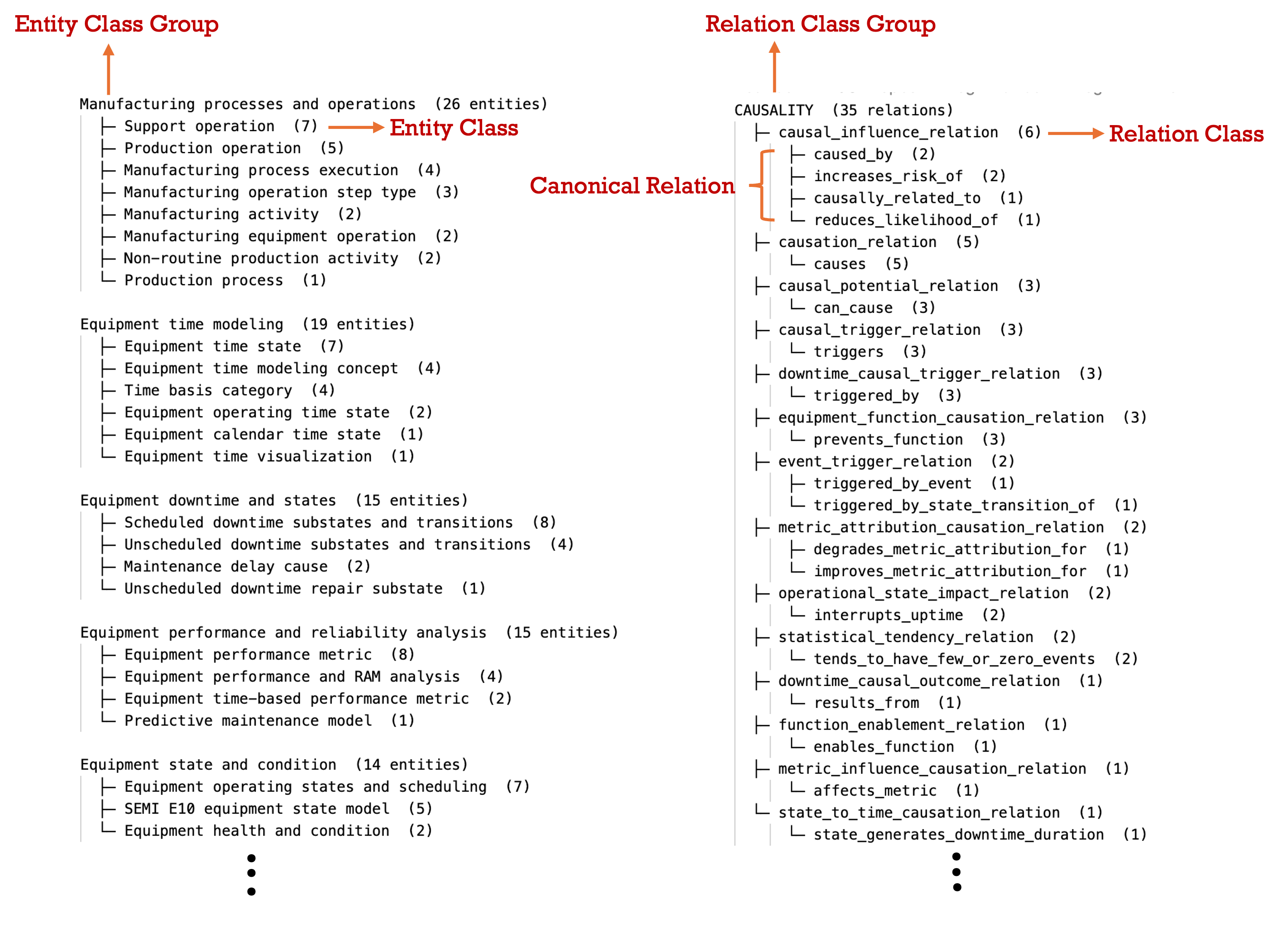}
    \caption{Induced entity and relation schema hierarchy for the case-study document.}
    \label{fig:case-schema-hierarchy}
\end{figure*}

\paragraph{What TRACE-KG recovers.}
% TRACE-KG converts the document into a consolidated, context-enriched graph while preserving direct links to source evidence. Figure~\ref{fig:case-wholekg} shows the resulting graph at document scale. Figure~\ref{fig:case-schema-hierarchy} shows that the induced schema is not flat: resolved entities are organized into interpretable groups such as manufacturing processes and operations, equipment time modeling, downtime and state semantics, performance and reliability analysis, and equipment condition/health. On the relation side, the induced schema is dominated by causal, trigger, and metric-impact patterns, reflecting the fact that RAM standards define performance through dependencies among states, events, and measurements.
TRACE-KG converts the document into a consolidated, context-enriched graph while preserving direct links to source evidence. Figure~\ref{fig:case-wholekg} shows the resulting graph at document scale. Figure~\ref{fig:case-schema-hierarchy} shows that the induced schema is not flat: resolved entities are organized into interpretable groups such as manufacturing processes and operations, equipment time modeling, downtime and state semantics, performance and reliability analysis, and equipment condition/health. Organizing technical knowledge into structured and interpretable forms is important for effective representation, transmission, and operational use of domain knowledge, and knowledge graphs provide a natural mechanism for supporting such organization \cite{esmaeili2022expounding}. On the relation side, the induced schema is dominated by causal, trigger, and metric-impact patterns, reflecting the fact that RAM standards define performance through dependencies among states, events, and measurements.

\begin{figure}[t]
    \centering
    \includegraphics[width=\columnwidth]{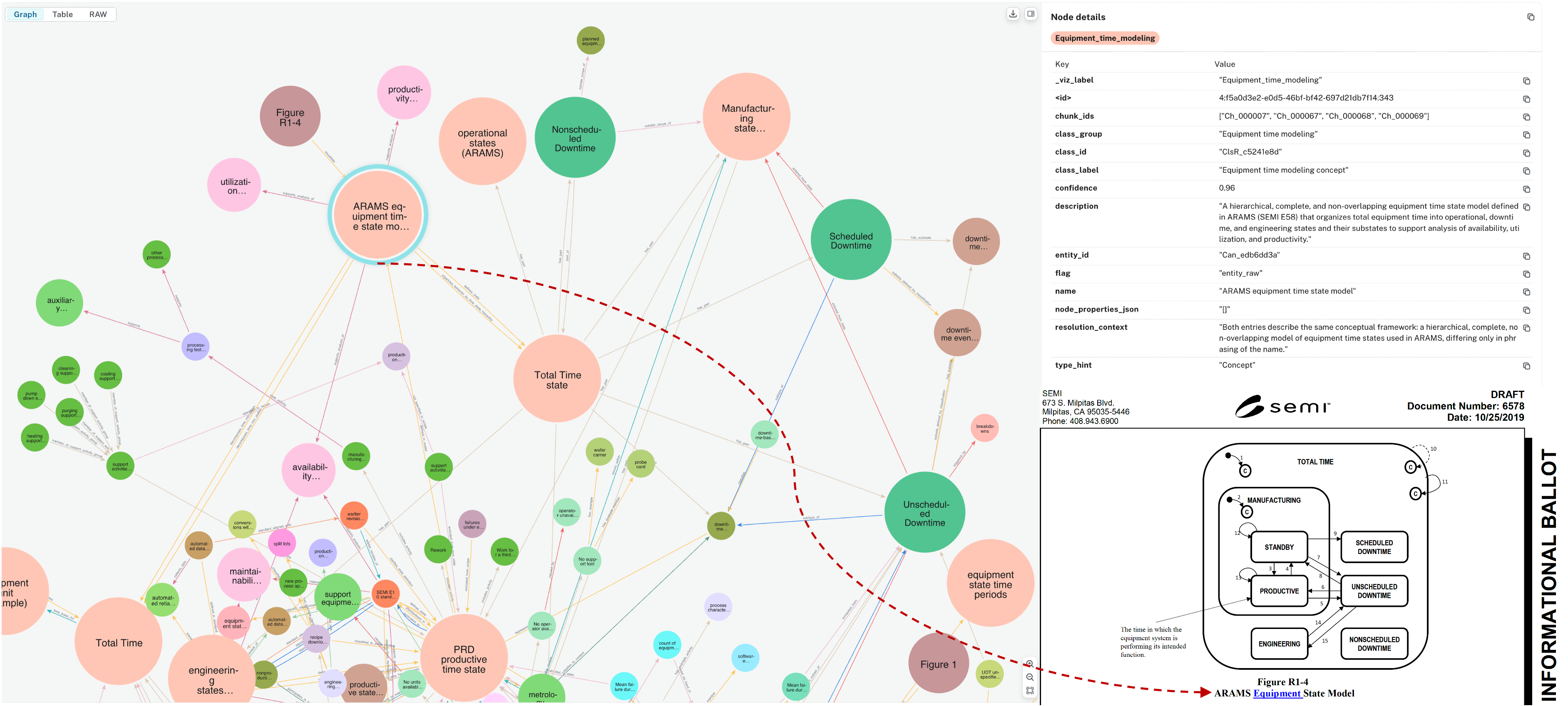}
    \caption{Traceability example (figure grounding): KG node linked to source diagram.}
    \label{fig:case-trace-figure}
\end{figure}

\begin{figure}[t]
    \centering
    \includegraphics[width=\columnwidth]{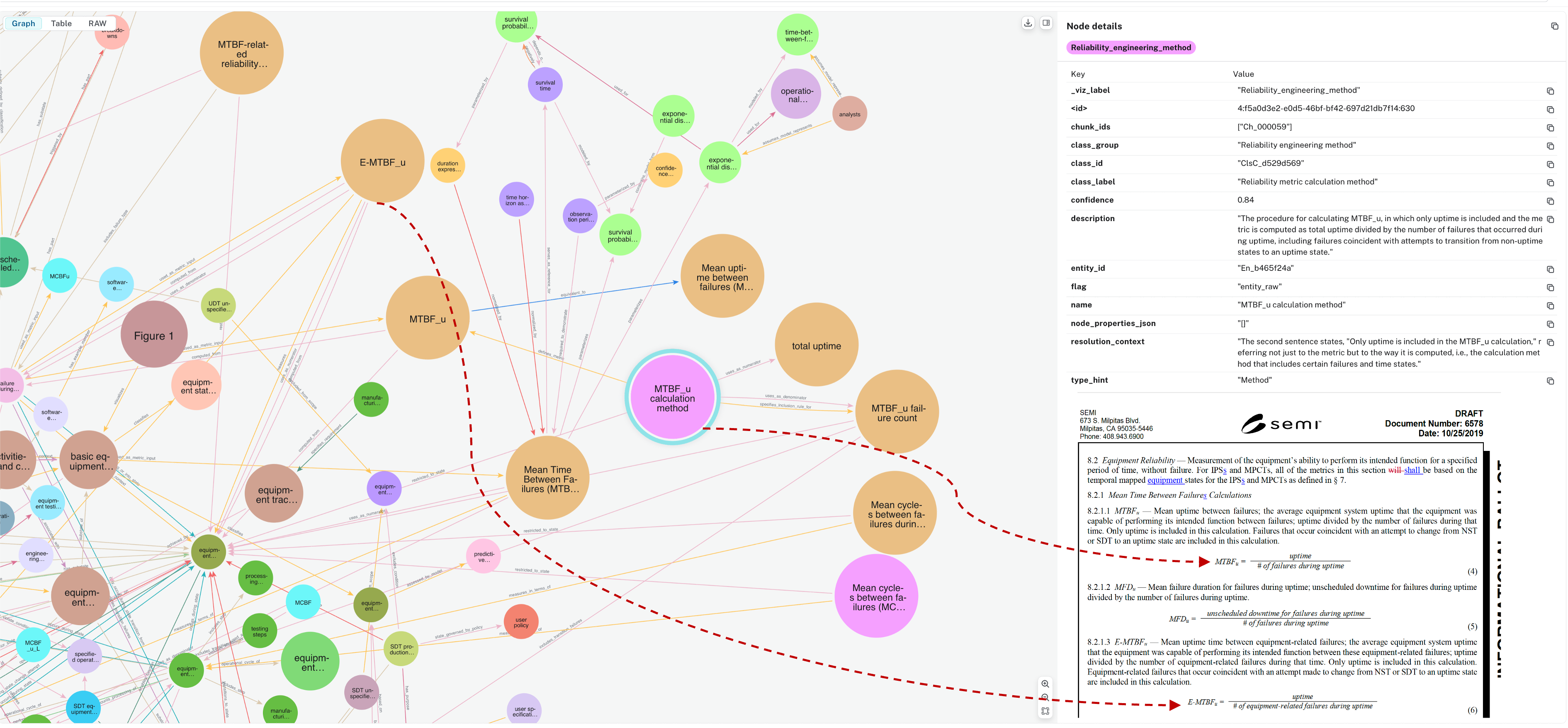}
    \caption{Traceability example (equation grounding): KG node linked to source equation.}
    \label{fig:case-trace-equation}
\end{figure}

\begin{figure}[t]
    \centering
    \includegraphics[width=\columnwidth]{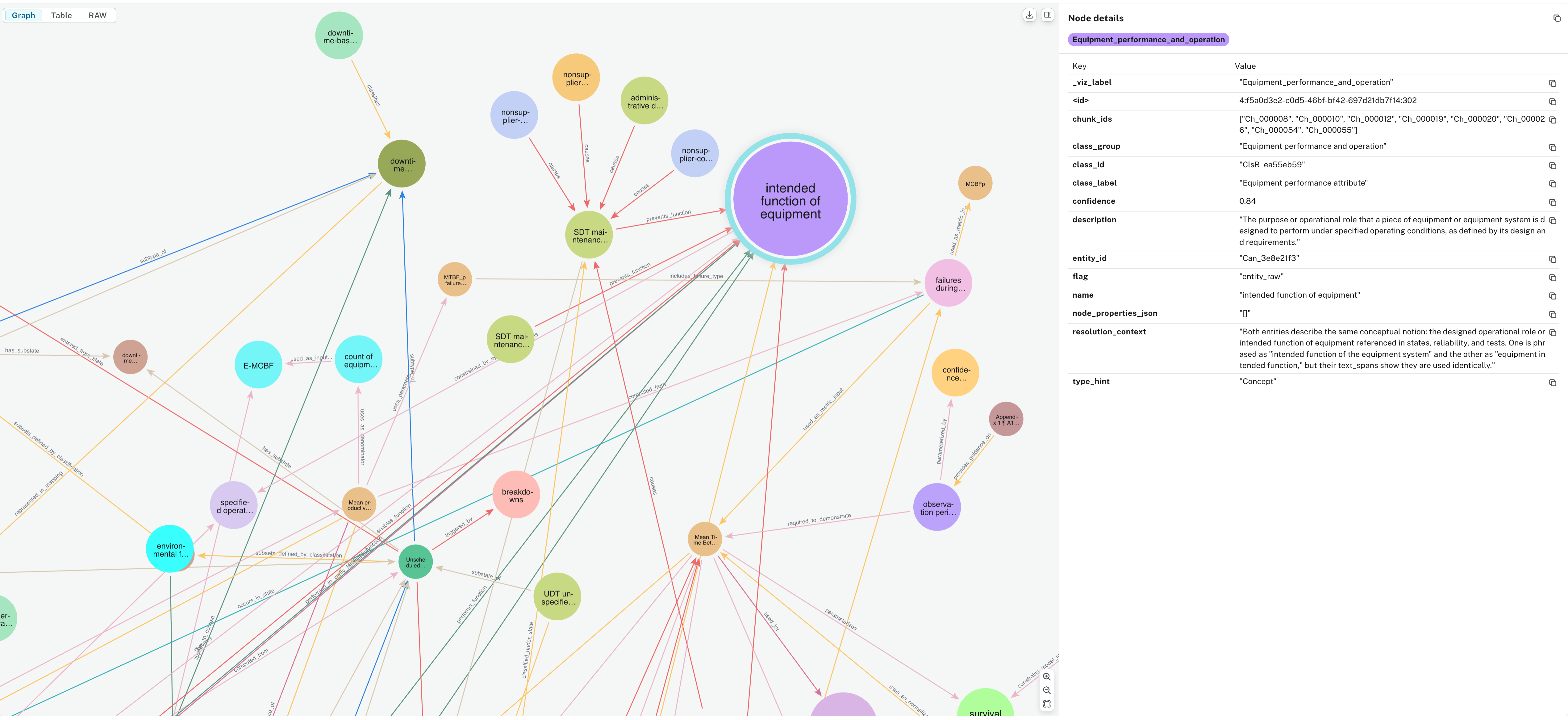}
    \caption{Example resolved entity with schema, confidence, and provenance.}
    \label{fig:case-node-example}
\end{figure}

\begin{figure}[t]
    \centering
    \includegraphics[width=\columnwidth]{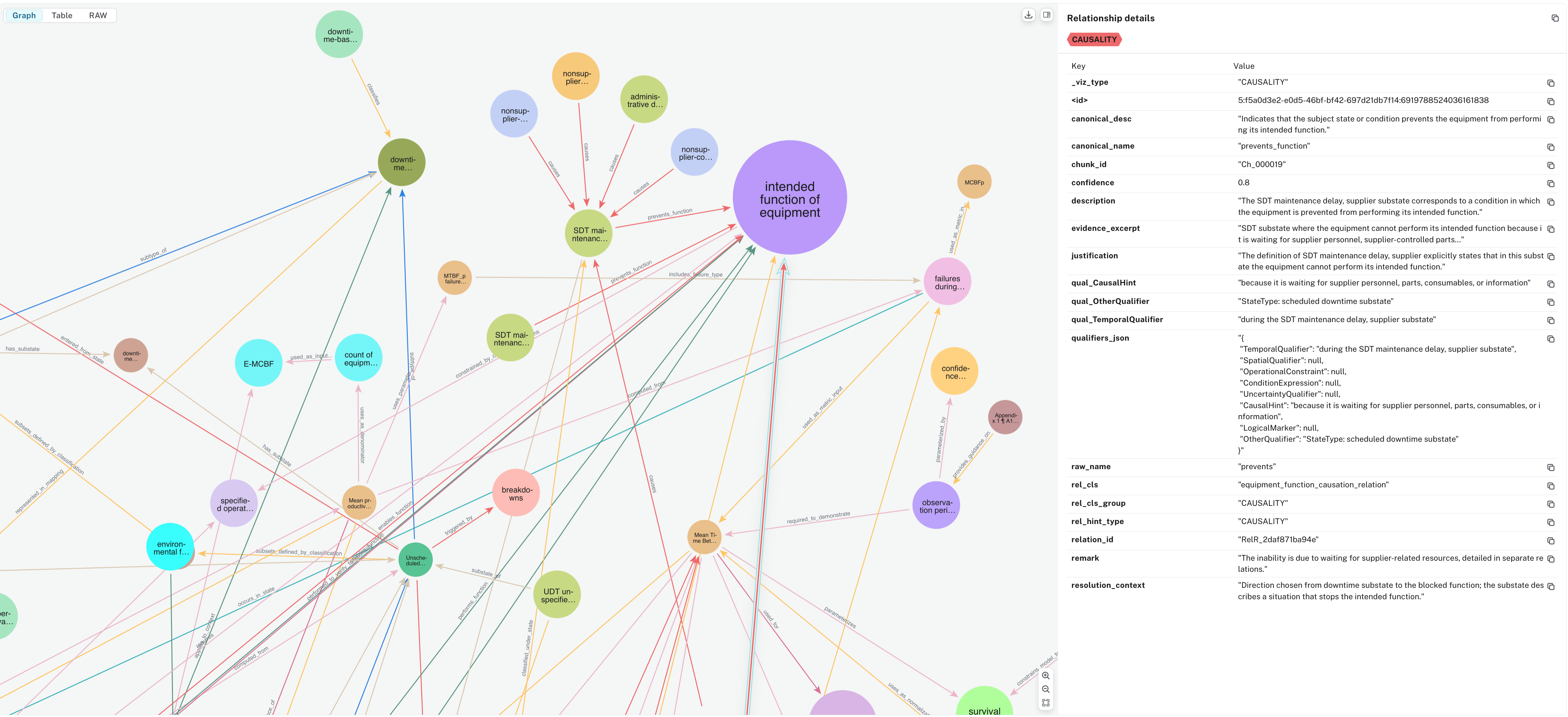}
    \caption{Example context-enriched relation with qualifiers and provenance.}
    \label{fig:case-rel-example}
\end{figure}

\paragraph{Traceability and inspection.}
% A key property of the case study is that the graph remains auditable against multimodal source material. Figure~\ref{fig:case-trace-figure} shows representative provenance view: the resolved ARAMS equipment time/state model is grounded back to its source diagram, and the resolved \(MTBF_u\) calculation method (shown in Figure ~\ref{fig:case-trace-equation}) is grounded back to its originating equation region. Figures~\ref{fig:case-node-example} and ~\ref{fig:case-rel-example} show inspected graph instances at the node and relation levels. The resolved node \emph{intended function of equipment} exposes induced class assignments, confidence, description, and supporting chunks, while the canonical relation \texttt{prevents\_function} exposes evidence, provenance, and structured qualifier fields. Together, these examples illustrate that TRACE-KG preserves not only a consolidated graph structure, but also the evidence needed for technical interpretation and review.
A key property of the case study is that the graph remains auditable against multimodal source material. This emphasis on evidence-grounded graph representations is consistent with prior work on explainable graph-based reasoning over heterogeneous sources, where entities and supporting evidence are linked explicitly to enable transparent and inspectable outputs \citep{christmann2023explainable}. Figure~\ref{fig:case-trace-figure} shows a representative provenance example: the resolved ARAMS equipment time/state model is grounded back to its source diagram, and the resolved \(MTBF_u\) calculation method (shown in Figure~\ref{fig:case-trace-equation}) is grounded back to its originating equation region. Figures~\ref{fig:case-node-example} and~\ref{fig:case-rel-example} show inspected graph instances at the node and relation levels. The resolved node \emph{intended function of equipment} exposes induced class assignments, confidence, description, and supporting chunks, while the canonical relation \texttt{prevents\_function} exposes evidence, provenance, and structured qualifier fields. Together, these examples illustrate that TRACE-KG preserves not only a consolidated graph structure, but also the evidence needed for technical interpretation and review.

\paragraph{Takeaway.}
This case study illustrates the type of document for which TRACE-KG is intended: standards-like technical material in which critical knowledge is distributed across text, equations, and diagrams, and in which analysts need a reusable semantic structure rather than isolated snippets. In this setting, the value of TRACE-KG lies not only in extracting entities and relations, but in organizing them into a coherent schema while maintaining end-to-end traceability.

\onecolumn

% =========================
% Figures (large + readable)
% =========================

\section{GenAI Usage Statement}
\label{sec:genai_usage}

Generative AI tools were used to assist with drafting and language refinement
of portions of this manuscript. Large language models were also accessed
programmatically through APIs as components of the TRACE-KG framework, as
described in the methodology and implementation details. All AI-assisted
content and outputs were reviewed and verified by the authors. The authors
take full responsibility for the final content of this work.

% \section{Notes}
% \subfile{sections/appendix-4-Notes}

% \clearpage
% % \twocolumn
% % \section{Algorithm pseudocode}
% % \subfile{sections/appendix-5-pseudoAlgorithm}

\end{document}